\documentclass[10pt,journal,compsoc,twoside]{IEEEtran}

\usepackage[nocompress]{cite}
\usepackage{graphicx}
\usepackage{amsmath}
\usepackage{amssymb}
\usepackage[caption=false,font=footnotesize,labelfont=sf,textfont=sf]{subfig}
\usepackage{url}
\usepackage[hidelinks]{hyperref}

\ifCLASSOPTIONcompsoc
 \usepackage[caption=false,font=footnotesize,labelfont=sf,textfont=sf]{subfig}
\else
 \usepackage[caption=false,font=footnotesize]{subfig}
\fi

\usepackage{algorithm}
\usepackage{algpseudocode}
\usepackage{enumitem}
\usepackage{mathtools}
\usepackage{siunitx}
\usepackage{multirow}

\DeclareMathOperator*{\argmax}{arg\,max}

\begin{document}

\title{Topology-Aware Non-Rigid\\ Point Cloud Registration}


\author{Konstantinos~Zampogiannis, Cornelia~Ferm\"uller, and~Yiannis~Aloimonos\IEEEcompsocitemizethanks{%
\IEEEcompsocthanksitem K. Zampogiannis (\texttt{kzampog@cs.umd.edu}) and Y. Aloimonos (\texttt{yiannis@cs.umd.edu}) are with the Department of Computer Science, University of Maryland, College Park, MD, 20742.
\IEEEcompsocthanksitem C. Ferm\"uller (\texttt{fer@umiacs.umd.edu}) is with the University of Maryland Institute for Advanced Computer Studies (UMIACS), University of Maryland, College Park, MD, 20742.}%
}

\IEEEtitleabstractindextext{%
\begin{abstract}
In this paper, we introduce a non-rigid registration pipeline for pairs of unorganized point clouds that may be topologically different.
Standard warp field estimation algorithms, even under robust, discontinuity-preserving regularization, tend to produce erratic motion estimates on boundaries associated with `close-to-open' topology changes.
We overcome this limitation by exploiting backward motion: in the opposite motion direction, a `close-to-open' event becomes `open-to-close', which is by default handled correctly.
At the core of our approach lies a general, topology-agnostic warp field estimation algorithm, similar to those employed in recently introduced dynamic reconstruction systems from RGB-D input.
We improve motion estimation on boundaries associated with topology changes in an efficient post-processing phase.
Based on both forward and (inverted) backward warp hypotheses, we explicitly detect regions of the deformed geometry that undergo topological changes by means of local deformation criteria and broadly classify them as `contacts' or `separations'.
Subsequently, the two motion hypotheses are seamlessly blended on a local basis, according to the type and proximity of detected events.
Our method achieves state-of-the-art motion estimation accuracy on the MPI Sintel dataset.
Experiments on a custom dataset with topological event annotations demonstrate the effectiveness of our pipeline in estimating motion on event boundaries, as well as promising performance in explicit topological event detection.
\end{abstract}

}

\maketitle

\IEEEdisplaynontitleabstractindextext

% \fi
\IEEEpeerreviewmaketitle

\IEEEraisesectionheading{\section{Introduction}\label{sec:introduction}}
\IEEEPARstart{M}{otion} estimation in 3D is a problem of great importance in computer vision, robotics, and computer graphics, playing a central role in a wide range of applications that include 3D scene reconstruction/modeling, human and object pose tracking, robot localization, augmented reality, human-computer interfaces and deformable shape manipulation.
The advent of affordable, commercial depth sensors has caused significant research effort on 3D motion estimation from 3D input, leading to the development of RGB-D algorithms for fast visual odometry \cite{huang2017visual,kerl2013robust}, efficient and accurate scene flow estimation \cite{jaimez2015primal,jaimez2017fast}, as well as notable SLAM systems for both static \cite{newcombe2011kinectfusion,whelan2015elasticfusion} and dynamic \cite{newcombe2015dynamicfusion,innmann2016volume} environments.

Given the availability of 3D input, dense non-rigid registration is
the most general motion estimation problem and it is particularly challenging.
In its general form, the problem can be described as computing a motion field, densely supported on the surface of a 3D shape, that deforms the latter in order to geometrically align it to another, fixed ``template'' shape.
This process of non-rigid 3D registration shares fundamental similarities with 2D image registration, known in the computer vision community as optical flow estimation: both problems pose similar challenges in deriving formulations that lead to accurate alignment while encoding reasonable prior constraints (regularization) to overcome ill-posedness.

A classical problem variant that is closely related to 3D non-rigid registration is that of RGB-D scene flow.
Given a pair of images, scene flow refers to the per-pixel 3D motion of observed points in space from the first frame to the second; optical flow refers to the per-pixel 2D projected motion.
There have been a number of successful recent works on scene flow estimation from RGB-D frame pairs, following both classical (variational) \cite{herbst2013rgb,quiroga2014dense,jaimez2015motion,jaimez2015primal,jaimez2017fast} and deep learning \cite{mayer2016large} frameworks.
While of great relevance to a number of motion reasoning tasks, RGB-D scene flow targets a specific instance of dense 3D motion estimation, as it inherently registers pairs of 2.5D geometries (depth maps).
This hinders its application in scenarios that require alignment of arbitrary 3D geometries, such as \emph{model-to-frame} registration for dynamic reconstruction or \emph{model-to-model} shape deformation.

Recently introduced dynamic reconstruction pipelines from RGB-D input \cite{newcombe2015dynamicfusion,innmann2016volume,dou2016acm,gao2018surfelwarp} solve a more general problem by implementing warp field optimization algorithms for their \emph{model-to-frame} registration step.
Despite adopting different approaches for their model representations and surface fusion steps, they all rely on similar, point cloud based formulations for non-rigid registration.
Scenes with dynamic topology are a challenging case for dynamic reconstruction systems: \cite{newcombe2015dynamicfusion} and \cite{innmann2016volume} make no provisions at all for these cases, while \cite{dou2016acm} and \cite{gao2018surfelwarp} deal with registration errors that occur because of dynamic topology at a subsequent stage, by discarding problematic regions and reinitializing model tracking.
The fully volumetric approaches of \cite{slavcheva2017killingfusion} and \cite{slavcheva2018sobolevfusion} do not use point representations for registration, directly aligning Signed Distance Fields (SDFs) \cite{curless1996volumetric} instead.
While they intrinsically handle topological changes, significant scalability limitations are introduced by relying on volumetric representations.
To the best of our knowledge, there exists no non-rigid \emph{point cloud} registration algorithm producing warp fields that are error-free on motion boundaries induced by dynamic scene topology.

We note that, throughout our discussion, we use the term `point cloud' to refer to a geometry representation by a \emph{discrete} point set sample of the underlying surface, as opposed to a volumetric 3D image.
Our broad definition does not preclude additional per-point attributes.
Therefore, oriented point clouds (point sets equipped with per-point normals) and surfel clouds (oriented point clouds with per-point radii) both fall within what we refer to simply as point cloud based representations.

\textbf{Contributions.} In this paper, we introduce a complete pipeline for the non-rigid registration of unorganized, oriented 3D point cloud pairs, which explicitly detects topology changes between the input point sets and produces piecewise-smooth warp fields that respect motion boundaries that result from these events.
At the core of our approach lies a general warp field estimation algorithm (Section \ref{sec:warp_field_estimation}), inspired by those employed in recent dynamic reconstruction systems from RGB-D input.
% We improve motion estimation on motion boundaries associated with topology changes in an efficient post-processing phase (Section \ref{sec:handling_topology_changes}) that relies on simple, intuitive tests of local deformation.
We improve motion estimation on motion boundaries associated with topology changes in an efficient post-processing phase (Section \ref{sec:handling_topology_changes}) that exploits the different properties of warp fields that are estimated in different directions (i.e. forward and backward) with respect to different types of topological events (i.e. `contact' or `separation', Section \ref{sec:motivation_overview}).
After explicitly detecting regions of topology change events by means of simple, intuitive tests of local deformation, our method blends the forward and inverted backward motion hypotheses on a local basis, based on the type and proximity of detected events, ensuring smooth, seamless hypothesis transitions on the deformed surface.
This stage makes no assumptions about the underlying registration engine and can be easily adapted for integration into existing pipelines.
The implementation of our warp field estimation module (without the topology event handling) is openly available as part of our point cloud processing library \cite{cilantro}.
Furthermore, the ability to detect and classify motion boundaries associated with dynamic topology is a byproduct of our pipeline that may be useful in tasks beyond geometric registration.
% As a byproduct that may be useful in tasks beyond geometric registration, our method extracts regions of the deformed geometry that are likely to lie on motion boundaries associated with topology changes, and broadly classifies them as either `contacts' or `separations' (Section \ref{sec:event_handling_evaluation}).

After discussing related work, we present our proposed method in detail in Section \ref{sec:approach}.
In Section \ref{sec:experiments}, we provide an extensive series of quantitative and qualitative experiments for the evaluation of our approach in terms of both registration accuracy and topological event handling.

\section{Related Work}\label{sec:related}
\textbf{RGB-D scene flow estimation.}
The term `scene flow' was introduced in \cite{vedula1999three} to refer to ``the  three-dimensional  motion field of points in the world, just as optical flow is the two-dimensional motion field of points in an image.''
Since then, significant research focus has shifted towards scene flow estimation from RGB-D input.
The formulation of \cite{herbst2013rgb} couples an $L^1$-norm data term derived from the \emph{optical flow} and \emph{range flow} \cite{spies2002range} constraints with weighted TV regularization.
In \cite{quiroga2014dense}, the authors follow a similar variational approach but use a rigid motion parameterization of the flow field, computing 6DoF per-pixel transformations and enforcing a local rigidity prior on the solution.
A 6DoF local parameterization is also used in \cite{hornacek2014sphereflow}, which introduces a correspondence search mechanism that relies on 3D spheres rather than image plane patches, and effectively handles large displacements.
In \cite{sun2015layered}, a probabilistic approach for joint segmentation and motion estimation method is proposed; a depth-based segmentation is used for motion estimation, which is in turn regularized based on the mean rigid motion of each layer.
A joint segmentation and scene flow estimation method is also presented in \cite{jaimez2015motion}, which assumes that the scene movement can be described by a small number of latent rigid motions.
Starting with a spatial $k$-means clustering for the motion label initialization, the algorithm iterates between motion estimation and segmentation (soft labeling), merging labels in the process.
In \cite{jaimez2015primal}, the first real-time RGB-D variational scene flow method is introduced, achieving state-of-the-art accuracy.
An efficient joint odometry and piecewise-rigid scene flow estimation method is proposed in \cite{jaimez2017fast}, where the scene is segmented into `static' and `moving' geometric clusters, from which odometry and independent non-rigid motions are computed.

As mentioned in our introduction, scene flow solves a somewhat restricted problem in the context of dense 3D registration, as the support of the computed motion field is image bound.

% \textbf{Shape manipulation.}
% ARAP \cite{sorkine2007rigid}
% ED \cite{sumner2007embedded}
% \textbf{Discontinuity preserving optical flow.}
% \cite{werlberger2009anisotropic}
% \cite{monzon2016robust}
% \cite{deriche1995optical}

% To further improve motion boundary preservation, in addition to using robust regularizers, several optical flow formulations have introduced isotropic or anisotropic weighting of the penalty terms.
% However, these adaptations are not directly applicable to a (non-volumetric) point cloud registration setting.

\textbf{Dynamic scene reconstruction.}
General non-rigid 3D registration algorithms have been developed in the context of online reconstruction of dynamic scenes from RGB-D input.
Most of them are formulated within a non-rigid Iterative Closest Point (ICP) framework, similar to the one introduced in \cite{amberg2007optimal}, with the goal of registering a point cloud representation of the scene model to the current frame, while there also exist purely volumetric approaches \cite{slavcheva2017killingfusion} that align Signed Distance Field (SDF) geometry representations.
DynamicFusion \cite{newcombe2015dynamicfusion} was the first system to achieve high quality, real-time dense reconstructions from RGB-D input.
While it performs volumetric (SDF) fusion \cite{curless1996volumetric}, its warp field estimation algorithm is based on oriented point cloud renderings of the model geometry.
The estimated warp field is defined on a sparse `Embedded Deformation' (ED) graph \cite{sumner2007embedded}, with a 6DoF transformation attached to each node, and its evaluation on arbitrary points is performed via interpolation.
The registration objective consists of a point-to-plane ICP cost, coupled with an `As-Rigid-As-Possible' (ARAP) \cite{sorkine2007rigid}, hierarchically defined regularization term, both under robust loss functions.
The non-rigid tracker of VolumeDeform \cite{innmann2016volume} does not rely on an ED graph and estimates individual 6DoF transformations for every source geometry point.
Its cost function consists of a dense point-to-plane cost, a sparse point-to-point term derived from SIFT \cite{lowe2004distinctive} correspondences, and an ARAP prior based on a `flat' neighborhood graph, with all terms being quadratic.
Fusion4D \cite{dou2016acm} combines the input of multiple range cameras for the task of dynamic reconstruction, using an ED warp field parameterization and following a similar registration objective formulation that additionally includes a `visual hull' term.
CoFusion \cite{ruenz2017icra} and MaskFusion \cite{runz2018maskfusion} segment, using semantic and motion cues, and reconstruct multiple moving objects in real-time, assuming that every object moves rigidly.
SurfelWarp \cite{gao2018surfelwarp} is a purely point (surfel) cloud based approach that also relies on an ED motion field representation and uses the same registration costs as DynamicFusion, but under the quadratic loss function.
On the other end of the spectrum, KillingFusion \cite{slavcheva2017killingfusion} and SobolevFusion \cite{slavcheva2018sobolevfusion} are purely volumetric approaches that rely on direct SDF-to-SDF alignment \cite{slavcheva2018sdf} via variational minimization under novel regularizers that enforce the motion field to be isometric and preserve level set geometry.

All of the above systems produce results of remarkable quality, especially given their real-time budget.
However, with the exceptions of \cite{dou2016acm}, \cite{gao2018surfelwarp}, \cite{slavcheva2017killingfusion}, and \cite{slavcheva2018sobolevfusion}, they cannot handle scenes with dynamic topology, with the `close-to-open' case (`separation', in our terminology) being particularly problematic.
According to our introductory discussion, \cite{dou2016acm} and \cite{gao2018surfelwarp} deal with these cases essentially by discarding affected regions, while the volumetric registration approaches of \cite{slavcheva2017killingfusion} and \cite{slavcheva2018sobolevfusion} are inherently immune to these events.
Our proposed method is the first to tackle dynamic topology within the context of \emph{motion estimation} and within a scalable \emph{point-based} representation framework.

\section{Our Approach}\label{sec:approach}

\subsection{Problem statement}\label{sec:problem_statement}
Given a pair of unorganized 3D point sets, our goal is to estimate a warping function that non-rigidly deforms the first point cloud (source geometry) towards the second one (target) in a piece-wise smooth manner.

Let $S=\left\{x_i^s\right\}\subset\mathbb{R}^3$ and $D=\left\{x_i^d\right\}\subset\mathbb{R}^3$ be the source and target geometry point sets, respectively, and $\mathcal{W}:\mathbb{R}^3\mapsto\mathbb{R}^3$ be a warping function.
In our non-rigid alignment setting, $\mathcal{W}$ is required to have the following properties:
\begin{itemize}[noitemsep,topsep=0pt]
    \item The image of point set $S$ via $\mathcal{W}$, $\mathcal{W}\left[S\right]$, should be aligned as close to the target geometry $D$ as possible.
    Typically, this is formulated as the minimization of the sum of residuals between points in $\mathcal{W}\left[S\right]$ and their corresponding points (e.g., nearest neighbors) in $D$.
    \item Local transformations of neighboring points that lie on the same moving surface in $S$ should be similar; i.e. $\mathcal{W}$ should be \emph{smooth}.
    At the same time, motion discontinuities should be preserved: neighboring points in $S$ that lie on independently moving surfaces should be allowed to have different local transformations.
    This combined prior is known as \emph{piecewise-smoothness}.
\end{itemize}
In a typical registration objective minimization formulation, the first property is expressed by the sum of registration residuals (e.g., point-to-point and/or point-to-plane distances) over corresponding point pairs in the objective (\emph{data term}), while the second one renders the otherwise under-constrained problem well-posed by introducing terms that penalize differences in local transformations of neighboring points (\emph{regularization term}).

The loss function used to model the regularization penalty terms, plays an important role in the behavior of the warping function in motion boundary regions.
For example, it is well known from the optical flow literature that quadratic regularization tends to oversmooth motion boundaries.
On the other hand, robust loss functions (e.g., $L^1$-norm approximations for the penalty terms) are more effective in producing piecewise-smooth motion fields that preserve discontinuities.

% To further improve motion boundary preservation, in addition to using robust regularizers, several optical flow formulations have introduced isotropic or anisotropic weighting of the penalty terms.
% However, these adaptations are not directly applicable to our (non-volumetric) point cloud registration setting.

In this work, we focus on estimating warp fields that respect motion boundaries resulting from changes in scene \emph{topology}.
Our notion of topology is directly derived from object-level connectivity: a change in scene topology can occur either when two or more separate objects come into contact or when two or more initially connected objects separate.
We note that we use the term `object' simply to refer to independently moving scene surface regions, without attaching to them any semantic meaning or assuming prior knowledge thereof.

In the following, we show that simply adopting a robust loss function for regularization still produces visible warping artifacts in motion discontinuity regions that result from scene topology changes, and we present a complete registration pipeline that effectively and efficiently solves this problem.

\begin{figure}[!t]
    \centering
    \subfloat[Left, middle: color frames (source and target) captured by an RGB-D camera. Right: warped source frame by the result of a standard non-rigid registration algorithm.]{\includegraphics[width=\columnwidth]{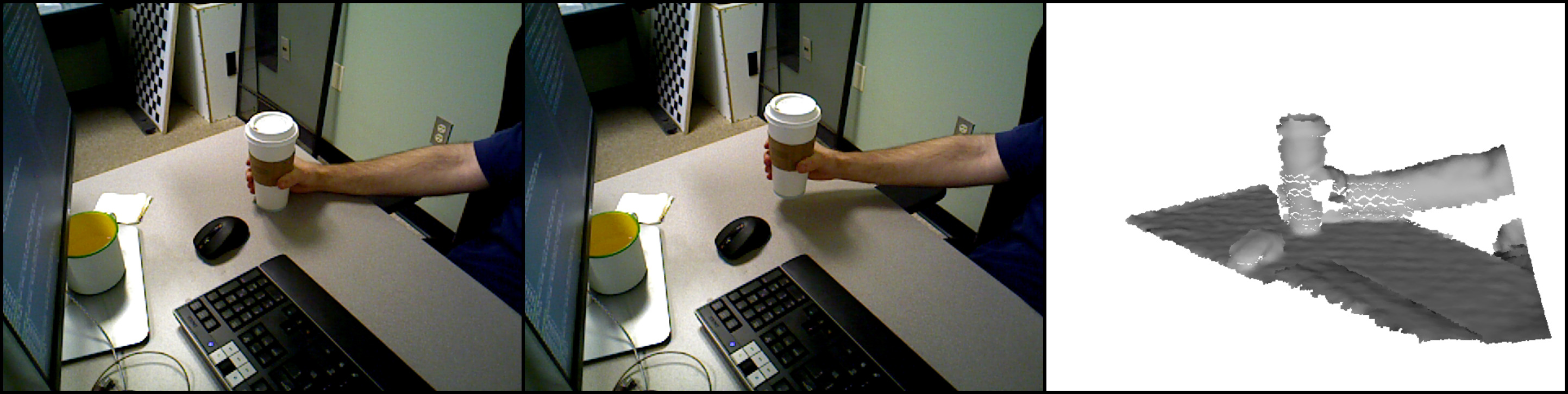}%
    \label{fig:intermediate_top}}
    \hfil
    \subfloat[Results of our proposed method. Left: detected (red) topological event regions (Section \ref{sec:topology_event_detection}). Middle: blending weight $w_b^i\in{[}0,1{]}$ for the inverted backward hypothesis using a `blue-to-red' colormap (blue: forward, red: inverted backward) shown at the bottom (Section \ref{sec:topology_hypothesis_blending}). Right: source frame transformed by our topology-aware warp field.]{\includegraphics[width=\columnwidth]{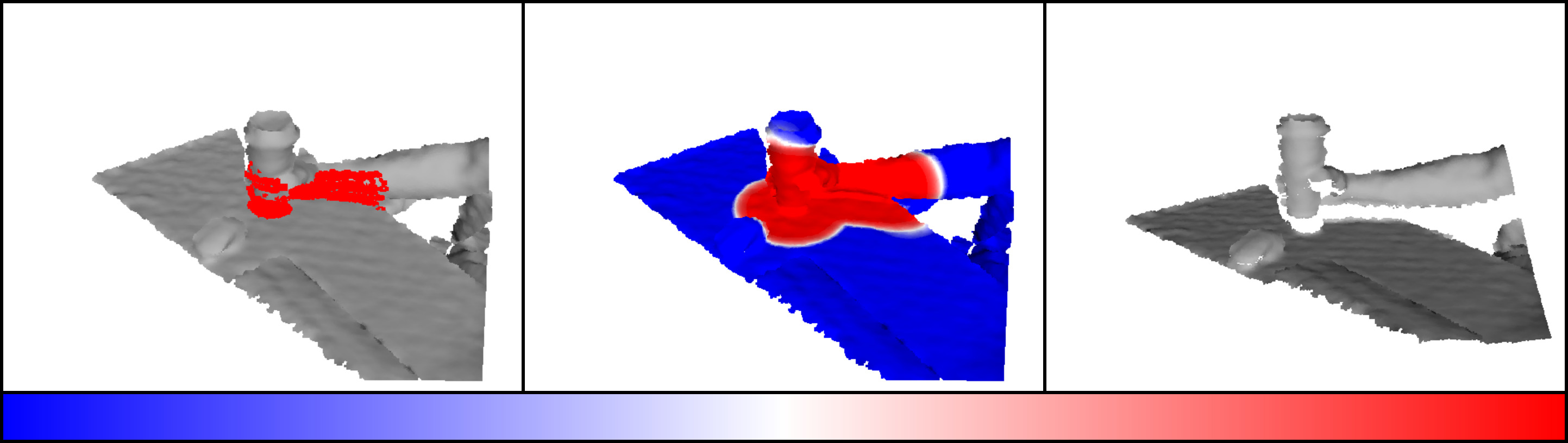}%
    \label{fig:intermediate_bot}}
    \caption{Non-rigid registration under a `close-to-open' topology change.}
    \label{fig:intermediate}
    % \vspace{-1em}
\end{figure}

\subsection{Motivation and overview of our approach}\label{sec:motivation_overview}
Motion estimation errors on motion boundaries typically manifest as oversmoothing of the warp field because of excessive regularization and can be suppressed by eliminating regularization penalty terms for points in $S$ that lie on different sides of the discontinuity.
However, without any knowledge about $S$ and its motion (e.g., some form of segmentation into independently moving objects), we cannot obtain a ``correct'' regularization graph a priori.
Instead, the common choice is to use a $k$-NN graph of points in $S$ to define the regularization terms.
It is easy to see that this choice is particularly problematic in cases where connected objects in $S$ move apart in $D$, as $k$-NN regularization over $S$ will introduce penalty terms that relate points that lie on different objects, resulting in some amount of motion field oversmoothing over the separation boundary.

Such a challenging scenario that involves object `separations' is depicted in Fig. \ref{fig:intermediate_top}, where the general, topology-agnostic warp field estimation algorithm described in Section \ref{sec:warp_field_estimation} was used to non-rigidly align two RGB-D frames.
Despite the fact that the algorithm's regularization term is formulated based on the robust, discontinuity-preserving Huber-$L^1$ loss function (see Section \ref{sec:setup_details} for parameter details), the registration result (warped source geometry) shows visible artifacts near the object separation areas.
Quadratic regularization is known to induce even more excessive smoothing on motion boundaries.
Since quadratic and $L^1$-norm approximation regularization types are the most commonly used ones in the literature, most current non-rigid registration algorithms are expected to exhibit a very similar behavior on these types of motion boundaries.

\begin{figure}[!t]
	\centering
	\includegraphics[width=0.98\columnwidth]{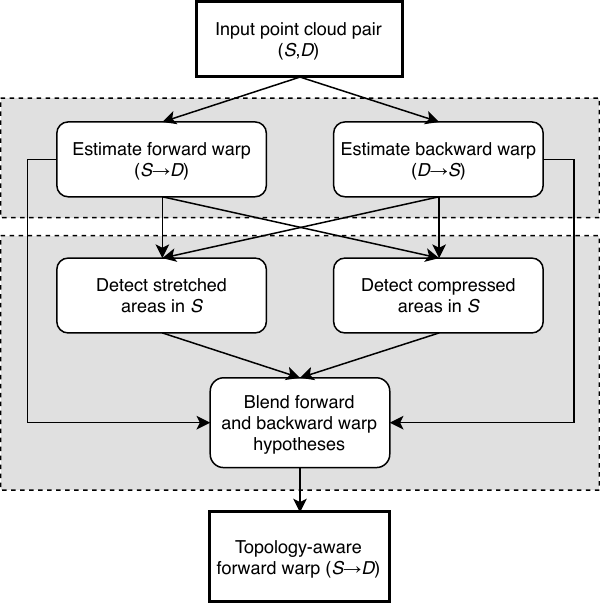}
    \caption{Overview of our topology-aware non-rigid registration pipeline.}
    \label{fig:overview}
    % \vspace{-1em}
\end{figure}

At the same time, it is clear that any change in topology between $S$ and $D$ will be directly reflected on the (different) nearest-neighbor graph structures of the source and target geometries.
We exploit this fact in the following way.
Consider the case where two or more objects that are connected in $S$ become separate in $D$.
As discussed above, estimating the warp field $\mathcal{W}_S^f$ that aligns $S$ to $D$ using the source geometry's $k$-NN graph to define the regularization penalties is expected to result in some amount of oversmoothing over the motion boundary of the separation.
However, in the \emph{backward} motion direction (from $D$ to $S$), the same topology change manifests as a connection of separate objects.
Estimating the backward warp field $\mathcal{W}_D^b$ that aligns $D$ to $S$ using the target ($D$) geometry's $k$-NN graph to define the regularization penalties should not exhibit any oversmoothing over the connection boundary, because the corresponding $k$-NN graph edges that would define regularization terms over the discontinuity are not there in the first place.
Inverting the warping function $\mathcal{W}_D^b$ yields another \emph{forward} warp field hypothesis, $\mathcal{W}_S^b$, that will be free of oversmoothing over separation motion boundaries.
Of course, the latter, being derived from $\mathcal{W}_D^b$, may suffer from oversmoothing over motion boundaries that correspond to object separations in the backward motion direction (from $D$ to $S$), or, equivalently, to objects coming into contact from $S$ to $D$.
These cases are expected to be handled correctly in the first place by the standard forward warp, $\mathcal{W}_S^f$.

Based on the above observations, the standard \emph{forward} warp $\mathcal{W}_S^f$ is expected to exhibit good behavior over \emph{contact} boundaries, but to oversmooth \emph{separation} boundaries.
On the other hand, the \emph{inverted backward} warp $\mathcal{W}_S^b$ is expected to behave the opposite way, preserving \emph{separation} discontinuities, but possibly blurring motion estimates in \emph{contact} areas.
Our proposed registration pipeline builds upon this idea by first detecting regions in $S$ that are likely to be contact or separation boundaries, and then locally blending the warp hypotheses $\mathcal{W}_S^f$ and $\mathcal{W}_S^b$ accordingly in a seamless manner.
The final result is a piecewise-smooth warp field that aligns $S$ to $D$ and respects motion boundaries because of changes in scene topology.

% We propose an efficient two-phase approach for estimating piecewise-smooth warp fields in a topology-aware manner.
% Our method builds on a general non-rigid registration algorithm, similar to the ones used in recent dynamic scene reconstruction pipelines, and handles motion estimation errors because of scene topology changes by explicitly detecting the latter in a post-processing phase.

An overview of our approach is provided in Fig. \ref{fig:overview}.
The (topology-agnostic) warp field estimation algorithm used to obtain the initial forward and backward warp hypotheses is described in detail in Section \ref{sec:warp_field_estimation}.
Our topology event detection mechanism, as well as our local hypothesis blending approach, are presented in Section \ref{sec:handling_topology_changes}.

\subsection{Warp field estimation}\label{sec:warp_field_estimation}
We implement our warp field estimation algorithm within a non-rigid Iterative Closest Point (ICP) framework \cite{amberg2007optimal}, similarly to the non-rigid trackers used in the recently introduced dynamic reconstruction pipelines of \cite{newcombe2015dynamicfusion}, \cite{innmann2016volume}, and \cite{gao2018surfelwarp}.

\begin{algorithm}[!t]
\caption{$\mathcal{W} = \textsc{NonRigidICP}(D, S, \mathcal{W}_0)$}
\label{alg:nr_icp}
\begin{algorithmic}[1]
\State $\mathcal{W}\gets \mathcal{W}_0$
\Repeat
    \State $S^\prime \gets \mathcal{W}[S]$
    \State $\mathcal{C} \gets \textsc{FindCorrespondences}(D, S^\prime)$
    \State $\mathcal{W}_{\textrm{iter}} \gets \textsc{OptimizeWarpField}(D, S^\prime, \mathcal{C})$
    \State $\mathcal{W} \gets \mathcal{W}_{\textrm{iter}}\circ \mathcal{W}$
\Until{$\mathcal{W}_{\textrm{iter}}$ is close to the identity warp}
\end{algorithmic}
\end{algorithm}

Given the source and target geometries $S$ and $D$, represented as oriented point clouds, as well as an initial estimate $\mathcal{W}_0$ of the unknown warp field $\mathcal{W}$ (usually taken as the identity warp), the algorithm iteratively refines the latter until convergence has been reached.
At the top level, the process iterates between a point correspondence search step between the warped source $\mathcal{W}[S]$ (according to the current $\mathcal{W}$ estimate) and $D$, and a warp field optimization step that updates $\mathcal{W}$ given the established point correspondences (Algorithm \ref{alg:nr_icp}).
The two algorithm phases are presented in detail in the following.

\subsubsection{Correspondence association}\label{sec:correspondence_association}
Our framework supports two complementary types of point correspondences between the (warped) source and the target geometries: \emph{dense} correspondences that are established based on spatial point proximity, and \emph{sparse} correspondences that result from keypoint matching.
Each individual correspondence is represented as a pair of point indices, whose first component indexes a point in $S$ and its second one a point in $D$: $\mathcal{C} = \left\{\mathcal{C}_\textrm{dense},\mathcal{C}_\textrm{sparse}\right\}$, where $\mathcal{C}_\textrm{dense},\mathcal{C}_\textrm{sparse}\subseteq\{1,\ldots,\lvert S\rvert\}\times\{1,\ldots,\lvert D\rvert\}$.

We support two mechanisms to establish dense correspondences.
By default, we assume that both $S$ and $D$ are arbitrary, unorganized point sets and we establish dense correspondences by finding the nearest-neighbor in $D$, in terms of Euclidean distance, of each point in $\mathcal{W}[S]$, with the search being performed efficiently by parallel $k$d-tree queries.
For certain applications that only require frame-to-frame (2.5D-to-2.5D) or model-to-frame (3D-to-2.5D) registration, we can further speed up the process by obtaining \emph{projective} correspondences.
This amounts to projecting $S$ and $D$ onto the target frame image and extracting correspondences based on points that are projected to the same pixel.
This is the mechanism adopted in most real-time reconstruction pipelines \cite{newcombe2011kinectfusion,newcombe2015dynamicfusion,innmann2016volume}.

In many common situations, dense geometric/depth correspondences alone are not enough to disambiguate the underlying motion.
For example, tracking points on flat surfaces that lack geometric texture and slide parallel to each other may exhibit drift.
Establishing robust keypoint correspondences between the source and target geometries can effectively mitigate this problem.
We assume that our input geometries are equipped with sparse interest points; our sparse correspondences are established by the interest point descriptor matches between $S$ and $D$.
In our implementation, we focus on input geometries that are either RGB-D frames or 3D reconstructions from RGB-D input.
The availability of regular images along with (registered) geometry allows us to adopt SIFT keypoint \cite{lowe2004distinctive} (lifted to 3D) matches for our sparse correspondences.

To make optimization more stable, we discard correspondence candidates from the above mechanisms that do not meet some basic proximity and local similarity criteria.
Let $\left\{n_i^s\right\}$, $\left\{n_i^d\right\}$, and $\left\{c_i^s\right\}$, $\left\{c_i^d\right\}$ be the surface normals and colors (e.g., RGB value 3-vectors) of the source and target geometries, indexed in the same way as their support points in $S$ and $D$.
A correspondence candidate $(i,j)\in\{1,\ldots,\lvert S\rvert\}\times\{1,\ldots,\lvert D\rvert\}$ is considered valid and used in the optimization if all of the following hold:
\begin{itemize}[noitemsep,topsep=0pt]
    \item $\lVert x_i^s-x_j^d\rVert_2 < \theta_d$
    % \item $\arccos{\left({n_i^s}^\top n_j^d\right)} < \theta_n$
    \item $\arccos{({n_i^s}^\top n_j^d)} < \theta_n$
    \item $\lVert c_i^s-c_j^d\rVert_2 < \theta_c$
\end{itemize}
In the above, $\theta_d$ is a point distance threshold, $\theta_n$ is a normal angle threshold, and $\theta_c$ is a color ``distance'' threshold.

\subsubsection{Warp field optimization}\label{sec:warp_field_optimization}
Given a set of dense and sparse point correspondences, we shall now describe our warp field optimization step.
Modeling the warp field using locally affine \cite{amberg2007optimal} or locally rigid \cite{newcombe2015dynamicfusion} transformations provides better motion estimation results than adopting a simple translational local model, due to more effective regularization.
In our approach, we adopt a locally rigid (6DoF) model.

Instead of computing a unique rigid transformation for each point in $S$, we use the more efficient embedded deformation graph representation \cite{sumner2007embedded} for the warp field $\mathcal{W}$, similarly to \cite{newcombe2015dynamicfusion} and \cite{dou2016acm}.
Let $\mathcal{G}=\left\{(g_i,\sigma_i,T_i)\right\}$ be the set of virtual deformation nodes, where $g_i\in\mathbb{R}^3$ is the position of the $i$th node, $\sigma_i$ is a radius parameter that controls the $i$th node's area of effect, and $T_i\in SE(3)$ is the 6DoF rigid transformation attached to the $i$th node.
The deformation node positions are obtained by downsampling the source geometry $S$ by means of a voxel grid of bin size $r_b$.
This allows us to use a uniform radius parameter, $\sigma_\textrm{def}$, for all deformation nodes, so that $\sigma_i=\sigma_\textrm{def}$, for $i=1,\ldots,\lvert\mathcal{G}\rvert$.
A reasonable choice that ensures sufficient area of effect overlap among neighboring nodes is $\sigma_\textrm{def} = r_b/2$.
% A reasonable choice for $\sigma_i$ that ensures sufficient area of effect overlap among neighboring nodes is $\sigma_i=\sigma_\textrm{def}\equiv r_b/2$, for $i=1,\ldots,\lvert\mathcal{G}\rvert$.
% A reasonable choice for $\sigma_i$ that ensures sufficient area of effect overlap among neighboring nodes is $\sigma_i=r_b/2$, for $i=1,\ldots,\lvert\mathcal{G}\rvert$.
%
As in \cite{innmann2016volume}, each local transformation $T_i$ is parameterized during optimization by a 6D vector $\theta_i$ of 3 Euler angles and 3 translational offsets.
The effect of the warp field $\mathcal{W}$, represented by $\mathcal{G}$, on a point $x\in\mathbb{R}^3$ is given by interpolating the local node deformations in the neighborhood of $x$.
Let $\mathcal{N}(x)\subseteq\{1,\ldots,\lvert\mathcal{G}\rvert\}$ be the indices of the $k$-nearest neighbors of $x$ in $\mathcal{G}$.
The local transformation parameter vector at $x$ is given by:
\begin{equation}
    \theta(x) \equiv \frac{\sum_{i\in\mathcal{N}(x)}w_i(x)\theta_i}{\sum_{i\in\mathcal{N}(x)}w_i(x)}\textrm{,}
    \label{eq:parameter_interpolation}
\end{equation}
where $w_{i}(x) = \textrm{exp}\left(-\left\lVert x-g_i\right\rVert_2^2/(2\sigma_i^2)\right)$.
The image of $x$ via $\mathcal{W}$ is then:
\begin{equation}
    \mathcal{W}(x) \equiv \textrm{Rot}(\theta(x))x + \textrm{Trans}(\theta(x))\textrm{,}
    % \mathcal{W}(x) \equiv \textrm{SE3}(\theta(x))[x^\top,1]^\top\textrm{,}
    \label{eq:warp_image}
\end{equation}
% where $\textrm{SE3}(\theta)$ maps our 6D parameterization to an $SE(3)$ transformation matrix.
where $\textrm{Rot}(\theta)$ and $\textrm{Trans}(\theta)$ extract the rotation matrix and translation vector from our 6D parameterization.

We note that the above 6D parameterization is only used within optimization (line 5 of Algorithm \ref{alg:nr_icp}) and that both the estimated \emph{incremental} warp $\mathcal{W}_\textrm{iter}$ and the final composite estimate $\mathcal{W}$ have their node transformations $T_i$ expressed in terms of $SE(3)$ transformation matrices.
The fact that we continuously warp $S$ and compute $\mathcal{W}_\textrm{iter}$ starting from the identity warp, combined with the  smoothness prior imposed on the warp field (shown below), allows us to overcome any problems associated with Euler angle parameterizations of rotation.

Our registration objective, as a function of the unknown warp field $\mathcal{W}$, which in the context of Algorithm \ref{alg:nr_icp} is the incremental warp $\mathcal{W}_\textrm{iter}$, and the point correspondences $\mathcal{C}$ between $S$ and $D$, which are fixed for this step, is formulated as:
\begin{equation}
    E\left(D,S,\mathcal{C},\mathcal{W}\right) = E_\textrm{data}\left(D,S,\mathcal{C},\mathcal{W}\right) + \lambda_\textrm{stiff}E_\textrm{stiff}\left(\mathcal{W}\right)\textrm{.}
    \label{eq:registration_objective}
\end{equation}
Our data term, $E_\textrm{data}\left(D,S,\mathcal{C},\mathcal{W}\right)$, is a weighted sum of a point-to-plane and a point-to-point ICP cost:
\begin{equation}
\begin{multlined}
    E_\textrm{data}\left(D,S,\mathcal{C},\mathcal{W}\right) = \\ \qquad\ \ \  E_\textrm{plane}\left(D,S,\mathcal{C},\mathcal{W}\right) + \lambda_\textrm{point}E_\textrm{point}\left(D,S,\mathcal{C},\mathcal{W}\right)\textrm{.}
\end{multlined}
\label{eq:data_term}
\end{equation}
Pure point-to-plane metric optimization generally converges faster and to better solutions than pure point-to-point \cite{rusinkiewicz2001efficient}, and is the standard trend in recent rigid \cite{newcombe2011kinectfusion,whelan2015elasticfusion} and non-rigid \cite{newcombe2015dynamicfusion,innmann2016volume,gao2018surfelwarp} registration pipelines.
However, as discussed in Section \ref{sec:correspondence_association}, integrating a point-to-point term for robust point matches into the registration cost can effectively disambiguate motion estimation in cases where surfaces that lack geometric texture slide parallel to each other.
Similarly to \cite{innmann2016volume}, we use our dense geometric correspondences $\mathcal{C}_\textrm{dense}$ to define our point-to-plane cost and our sparse keypoint correspondences $\mathcal{C}_\textrm{sparse}$ for our point-to-point cost:
\begin{align}
    E_\textrm{plane}\left(D,S,\mathcal{C},\mathcal{W}\right)\ \ & =\ \ \  \sum_{\mathclap{(i,j)\in\mathcal{C}_\textrm{dense}}}\left({n_j^d}^\top\left(\mathcal{W}\left(x_i^s\right)-x_j^d\right)\right)^2\textrm{,}\label{eq:plane_term}\\
    E_\textrm{point}\left(D,S,\mathcal{C},\mathcal{W}\right)\ \ & =\ \ \  \sum_{\mathclap{(i,j)\in\mathcal{C}_\textrm{sparse}}}\left\lVert\mathcal{W}\left(x_i^s\right)-x_j^d\right\rVert_2^2\textrm{.}\label{eq:point_term}
\end{align}
Our regularization term $E_\textrm{stiff}\left(\mathcal{W}\right)$ directly penalizes differences in transformation parameters of neighboring virtual nodes of $\mathcal{G}$ under the robust Huber-$L^1$ loss function.
If $\mathcal{N}(i)\subseteq\{1,\ldots,\lvert\mathcal{G}\rvert\}$ is the set of indices of the $k$-nearest neighbors of $g_i$ in $\mathcal{G}$, our regularization term is formulated as:
\begin{equation}
    E_\textrm{stiff}\left(\mathcal{W}\right) = \sum_{i=1}^{\left\lvert\mathcal{G}\right\rvert}\sum_{j\in \mathcal{N}(i)}w_{ij}\psi_\delta\left( \theta_i - \theta_j\right)\textrm{,}
    \label{eq:regularization_term}
\end{equation}
where $w_{ij} = \textrm{exp}\left(-\left\lVert g_i-g_j\right\rVert^2/(2\sigma_{\textrm{def}}^2)\right)$ weights the pairwise penalties based on node distance, and $\psi_\delta(\Delta\theta)$ denotes the sum of the Huber loss function values over the 6 parameter residual components.
Parameter $\delta$ controls the point at which the loss function behavior switches from quadratic (squared $L^2$-norm) to absolute-linear ($L^1$-norm).
Since $L^1$-norm regularization is known to better preserve solution discontinuities, we use a small value of $\delta=10^{-4}$.
Given our locally rigid motion model, this regularization scheme enforces an `As-Rigid-As-Possible' \cite{sorkine2007rigid} prior to the estimated warp field.

The registration objective of equation \eqref{eq:registration_objective} is non-linear in the $6\mathcal\lvert G\rvert$ unknowns.
We minimize $E\left(D,S,\mathcal{C},\mathcal{W}\right)$ by performing a small number of Gauss-Newton iterations.
As in \cite{dou2016acm}, we handle non-quadratic terms using the square-rooting technique of \cite{engels2006bundle}.
At every step, we linearize $E$ around the current solution $\theta\in\mathbb{R}^{6\mathcal\lvert G\rvert}$ (vector concatenation of all node transformation parameters $\theta_i$) and obtain a solution increment $\hat{\theta}$ by solving the system of normal equations $J^\top J\hat{\theta}=J^\top r$, where $J$ is the Jacobian matrix of the residual terms in $E$ and $r$ is the vector of (negative) residual values.
We solve this sparse system iteratively, using the Conjugate Gradient algorithm with a diagonal preconditioner.

\subsection{Handling topology changes}\label{sec:handling_topology_changes}
In our post-processing phase for handling topology change motion boundaries, we first explicitly detect likely \emph{contact} and \emph{separation} regions in the source geometry, and proceed by appropriately blending our (default) \emph{forward} and \emph{inverted backward} warp hypotheses in a local yet seamless manner.
It follows from the discussion in Section \ref{sec:motivation_overview} that, as far as topology changes are concerned, the forward warp is only problematic in separation areas.
However, instead of only focusing on and amending separations, it is beneficial to also explicitly consider contact events.
As will become clear in the following, considering both types of events and treating them symmetrically robustifies their detection and allows for a less biased hypothesis blending scheme.

% \subsubsection{Preliminaries}\label{sec:topology_preliminaries}
Using the same notation as in Section \ref{sec:motivation_overview}, let $\mathcal{W}_S^f$ be the warp field that aligns $S$ to $D$ and $\mathcal{W}_D^b$ the backward warp that aligns $D$ to $S$, both computed by Algorithm \ref{alg:nr_icp}.
We will be using the `S' subscript for forward ($S\rightarrow D$) motion entities and the `D' subscript for backward ($D\rightarrow S$) ones.
For the needs of the following discussion, we will consider these motion fields to be represented by the \emph{per-point} local rigid transformations of their support geometries, so that:
\begin{align}
    \mathcal{W}_S^f &= \{{T_S^f}_i,\ i=1,\ldots,\lvert S\rvert\}\textrm{,}\label{eq:forward_warp_1}\\
    \mathcal{W}_D^b &= \{{T_D^b}_i,\ i=1,\ldots,\lvert D\rvert\}\textrm{.}\label{eq:backward_warp_1}
\end{align}
To invert $\mathcal{W}_D^b$ in order to obtain an alternative forward warp, we appropriately rebase its inverse local transformations on $S$.
To that end, we first compute the target geometry's image $\mathcal{W}_D^b[D]$ (which should be closely aligned to $S$) and, to each point $x_i^s\in S$, we assign the transformation ${T_D^b}_j^{-1}$, where $j$ is the index of the nearest neighbor of $x_i^s$ in the point set $\mathcal{W}_D^b[D]$.
Of course, we assume that the latter is indexed in the same way as $D$.
The inverted backward warp is then represented as:
\begin{equation}
    \mathcal{W}_S^b = \{{T_S^b}_i,\ i=1,\ldots,\lvert S\rvert\}\textrm{,}\label{eq:forward_warp_2}
\end{equation}
where ${T_S^b}_i={T_D^b}_j^{-1}$ and $j$ is the nearest neighbor index of $x_i^s$ in $\mathcal{W}_D^b[D]$.
Analogously, we obtain an alternative backward warp, by inverting our forward hypothesis:
\begin{equation}
    \mathcal{W}_D^f = \{{T_D^f}_i,\ i=1,\ldots,\lvert D\rvert\}\textrm{,}\label{eq:backward_warp_2}
\end{equation}
where ${T_D^f}_i={T_S^f}_j^{-1}$ and $j$ is the nearest neighbor index of $x_i^d$ in $\mathcal{W}_S^f[S]$.
To summarize, we have two forward motion ($S\rightarrow D$) warp hypotheses ($\mathcal{W}_S^f$ and $\mathcal{W}_S^b$) and two backward motion ($D\rightarrow S$) ones ($\mathcal{W}_D^b$ and $\mathcal{W}_D^f$).

\renewcommand{\algorithmicrequire}{\textbf{Input:}}
\renewcommand{\algorithmicensure}{\textbf{Output:}}
\begin{algorithm}[!t]
\caption{$\textsc{ExtractTopologyEvents}$}
\label{alg:extract_topology_events}
\begin{algorithmic}[1]
\Require{$S,\textrm{\textsc{Stretch}}_S^f,\textrm{\textsc{Stretch}}_S^b,\textrm{\textsc{Compress}}_S^f,\textrm{\textsc{Compress}}_S^b$}
\Ensure{$\textsc{Con},\textsc{Sep}$}
\vspace{0.3em}
\State $\textsc{Con} \gets \varnothing$
\State $\textsc{Sep} \gets \varnothing$
\For{$i = 1,\ldots,\lvert S\rvert$}
    \vspace{0.3em}
    \State $\textrm{stretch}\gets\max\left\{\textrm{\textsc{Stretch}}_S^f(i),\textrm{\textsc{Stretch}}_S^b(i)\right\}$
    \vspace{0.3em}
    \State $\textrm{compress}\gets\max\left\{\textrm{\textsc{Compress}}_S^f(i),\textrm{\textsc{Compress}}_S^b(i)\right\}$
    \vspace{0.5em}
    \If{$\textrm{stretch}>\tau\textrm{ \textbf{and} }\textrm{stretch}>\alpha\cdot\textrm{compress}$}
        \State $\textsc{Sep}\gets \textsc{Sep}\cup\left\{x_i^s\right\}$
    \EndIf
    \If{$\textrm{compress}>\tau\textrm{ \textbf{and} }\textrm{compress}>\alpha\cdot\textrm{stretch}$}
        \State $\textsc{Con}\gets \textsc{Con}\cup\left\{x_i^s\right\}$
    \EndIf
\EndFor
\end{algorithmic}
\end{algorithm}

\subsubsection{Detecting topology change events}\label{sec:topology_event_detection}
We detect topology change regions in $S$ based on how our warp estimates affect local neighborhoods of the source geometry.

Naturally, we expect that if $x_i^s\in S$ is close to a separation boundary, its distance to some of its neighbors in $S$ should increase after applying the correct warp to $S$.
We shall refer to this effect as neighborhood \emph{stretching}.
The dual case of a contact event manifests exactly the same way in the backward motion direction (stretching of neighborhoods of $D$), in which the event is perceived as a separation.
In the following, we will use a local measure of stretch over points in $S$ to detect separation areas, and map the same measure over $D$ in the backward direction onto $S$ to obtain a dual measure of ``compression'' that will allow us to detect contacts.

We quantify the above intuition by defining a local ``stretch'' operator for point $x_i\in X\subseteq\mathbb{R}^3$ under the warp $\mathcal{W}$ as the maximum ratio of the distance to its neighbors before and after applying $\mathcal{W}$:
\begin{equation}
    \textrm{\textbf{Stretch}}\left(i,X,\mathcal{W}\right) \equiv \max_{j\in\mathcal{N}(i)}\frac{\left\lVert \mathcal{W}(x_i)-\mathcal{W}(x_j) \right\rVert_2}{\left\lVert x_i-x_j \right\rVert_2}\textrm{,}
    \label{eq:stretch_operator}
\end{equation}
where $\mathcal{N}(i)\subseteq \{1,\ldots,\lvert X\rvert\}$ indexes the neighbors of $x_i$ in $X$ that lie within $\rho_s$ distance from it.
The choice of the neighborhood radius value $\rho_s$ depends on the scale and resolution of the input geometries.
For close-range point clouds acquired with Kinect-like cameras, we use $\rho_s=1.5\si{\centi\metre}$.

To each point in $S$, we associate one stretch value for each of our two forward warp hypotheses, according to definition \eqref{eq:stretch_operator}.
For $i=1,\ldots,\lvert S\rvert$, we have:
\begin{align}
    \textrm{\textsc{Stretch}}_S^f(i) &= \textrm{\textbf{Stretch}}\left(i,S,\mathcal{W}_S^f\right)\textrm{, and}\\
    \textrm{\textsc{Stretch}}_S^b(i) &= \textrm{\textbf{Stretch}}\left(i,S,\mathcal{W}_S^b\right)\textrm{.}
\end{align}
We also compute the local stretch of the target geometry $D$ under each of the two backward warps:
\begin{align}
    \textrm{\textsc{Stretch}}_D^f(i) &= \textrm{\textbf{Stretch}}\left(i,D,\mathcal{W}_D^f\right)\textrm{, and}\\
    \textrm{\textsc{Stretch}}_D^b(i) &= \textrm{\textbf{Stretch}}\left(i,D,\mathcal{W}_D^b\right)\textrm{,}
\end{align}
which we subsequently map onto $S$, interpreting them as a compression measure (contact indicator), according to:
\begin{align}
    \textrm{\textsc{Compress}}_S^f(i) &= \textrm{\textsc{Stretch}}_D^f\left( \textrm{NN}\left(\mathcal{W}_S^f\left(x_i^s\right),D\right) \right)\textrm{,}\\
    \textrm{\textsc{Compress}}_S^b(i) &= \textrm{\textsc{Stretch}}_D^b\left( \textrm{NN}\left(\mathcal{W}_S^b\left(x_i^s\right),D\right) \right)\textrm{,}
\end{align}
where $\textrm{NN}(x,X)\in \{1,\ldots,\lvert X\rvert\}$ is the index of the nearest neighbor of point $x$ in point set $X$.

Using the above point-wise stretch/compress values on $S$, we extract subsets of the source geometry that are likely to lie on topology change motion boundaries.
Let $\textsc{Sep}, \textsc{Con}\subseteq S$ be the sets of candidate separation and contact boundary points, respectively.
According to the above discussion, points on a separation boundary are expected to have high stretch scores, while points on a contact boundary should exhibit high local compression.
To decide whether a point in $S$  is a boundary candidate, we perform two symmetric tests per case that rely on two threshold values, an absolute score threshold $\tau$, and a relative (ratio) threshold $\alpha$.
A point of $S$ is a member of $\textsc{Sep}$ ($\textsc{Con}$) if and only if the maximum of its two stretch (compress) scores is greater than $\tau$ and also greater than $\alpha$ times its maximum compress (stretch) score.
The process is summarized in Algorithm \ref{alg:extract_topology_events}.
A sample output is shown in Fig. \ref{fig:intermediate_bot} (left), marked in red; note that $\textsc{Con}=\varnothing$ in this case.

As we will show in Section \ref{sec:event_handling_evaluation}, the above procedure is very effective at detecting and classifying topology changes, but, because of the continuous nature of our local stretch/compression measures and depending on the selected threshold values, it may produce ``false positives'' (e.g., in areas of actual deforming surface stretching or compression but constant topology).
However, under the assumption that our two forward warp hypotheses behave similarly in the false positive areas and, as will become clear in the next section, this does not affect our final warp estimate.

\subsubsection{Local hypothesis blending}\label{sec:topology_hypothesis_blending}
Our blending scheme produces a topology-aware warp field $\mathcal{W}_S$ by combining the forward warp hypotheses $\mathcal{W}_S^f$ and $\mathcal{W}_S^b$ on a per-point basis.
Our objective is to assign a higher weight to $\mathcal{W}_S^b$ (inverted backward warp) near separation areas, and ensure that $\mathcal{W}_S^f$ (forward warp) has a stronger weight near contact areas.
At the same time, it is desirable that point weights vary smoothly on $S$, so that our warp blending does not introduce seam artifacts on $\mathcal{W}_S[S]$ due to differences in our original warp hypotheses.

\begin{algorithm}[!t]
\caption{$\textsc{LocalHypothesisBlending}$}
\label{alg:local_hypothesis_blending}
\begin{algorithmic}[1]
\Require{$S,\mathcal{W}_S^f,\mathcal{W}_S^b,\textsc{Con}=\{c_i\},\textsc{Sep}=\{s_i\},\rho_e$}
\Ensure{$\mathcal{W}_S=\left\{{T_S}_i,\ i=1,\ldots,\lvert S\rvert \right\}$}
\State $\sigma \gets \rho_e/3$
\For{$i = 1,\ldots,\lvert S\rvert$}
    \State $\mathcal{N} \gets \textsc{RadiusSearch}\left(x_i^s,\textsc{Con},\rho_e\right)$
    \State $w_f\gets 1$
    \For{$j\in\mathcal{N}$}
        \State $w_f\gets w_f + \exp\left(-\left\lVert x_i^s-c_j\right\rVert^2/(2\sigma^2)\right)$
    \EndFor
    \State $\mathcal{N} \gets \textsc{RadiusSearch}\left(x_i^s,\textsc{Sep},\rho_e\right)$
    \State $w_b\gets 0$
    \For{$j\in\mathcal{N}$}
        \State $w_b\gets w_b + \exp\left(-\left\lVert x_i^s-s_j\right\rVert^2/(2\sigma^2)\right)$
    \EndFor
    \State $w\gets w_f + w_b$
    \State $w_f\gets w_f/w$
    \State $w_b\gets w_b/w$
    \State ${T_S}_i \gets \textrm{SE3}\left(w_f{T_S^f}_i + w_b{T_S^b}_i\right)$
\EndFor
\end{algorithmic}
\end{algorithm}

\renewcommand{\arraystretch}{1.2}
\begin{table*}[!t]
\caption{Evaluation on the MPI Sintel Dataset}
\label{tab:sintel_evaluation}
\vspace{-0.7em}
\centering
\begin{tabular}{| c | r r | r r | r r || r r | r r | r r |}
%
% \cline{2-13}
% \multicolumn{1}{c|}{} & \multicolumn{6}{c||}{\textbf{Frame-wise EPE over entire sequence}} & \multicolumn{6}{c|}{\textbf{Frame-wise EPE over entire sequence}}\\
% \cline{2-13}
% \multicolumn{1}{c|}{} & \multicolumn{2}{c|}{\textbf{PD-Flow}} & \multicolumn{2}{c|}{\textbf{F-Warp}} & \multicolumn{2}{c||}{\textbf{FB-Warp}} & \multicolumn{2}{c|}{\textbf{PD-Flow}} & \multicolumn{2}{c|}{\textbf{F-Warp}} & \multicolumn{2}{c|}{\textbf{FB-Warp}}\\
% \hline
% \multicolumn{1}{|c|}{\textbf{Sequence}} & \multicolumn{1}{c}{Median} & \multicolumn{1}{c|}{Mean} & \multicolumn{1}{c}{Median} & \multicolumn{1}{c|}{Mean} & \multicolumn{1}{c}{Median} & \multicolumn{1}{c||}{Mean} & \multicolumn{1}{c}{Median} & \multicolumn{1}{c|}{Mean} & \multicolumn{1}{c}{Median} & \multicolumn{1}{c|}{Mean} & \multicolumn{1}{c}{Median} & \multicolumn{1}{c|}{Mean}\\
% 
\hline
\multicolumn{1}{|c|}{\multirow{3}{*}{\textbf{Sequence}}} & \multicolumn{6}{c||}{\textbf{Frame-wise EPE over entire sequence}} & \multicolumn{6}{c|}{\textbf{Frame-wise AE over entire sequence}}\\
% \multicolumn{1}{c|}{} & \multicolumn{6}{c||}{\textbf{EPE}} & \multicolumn{6}{c|}{\textbf{AE}}\\
\cline{2-13}
\multicolumn{1}{|c|}{} & \multicolumn{2}{c|}{\textbf{PD-Flow}} & \multicolumn{2}{c|}{\textbf{F-Warp}} & \multicolumn{2}{c||}{\textbf{FB-Warp}} & \multicolumn{2}{c|}{\textbf{PD-Flow}} & \multicolumn{2}{c|}{\textbf{F-Warp}} & \multicolumn{2}{c|}{\textbf{FB-Warp}}\\
\cline{2-13}
\multicolumn{1}{|c|}{} & \multicolumn{1}{c}{Median} & \multicolumn{1}{c|}{Mean} & \multicolumn{1}{c}{Median} & \multicolumn{1}{c|}{Mean} & \multicolumn{1}{c}{Median} & \multicolumn{1}{c||}{Mean} & \multicolumn{1}{c}{Median} & \multicolumn{1}{c|}{Mean} & \multicolumn{1}{c}{Median} & \multicolumn{1}{c|}{Mean} & \multicolumn{1}{c}{Median} & \multicolumn{1}{c|}{Mean}\\
\hline
\texttt{alley\_1}    &    0.774    &    0.964    &    0.485    &    0.519    &    \textbf{0.485}    &    \textbf{0.516}    &    \textbf{7.005}    &    \textbf{7.319}    &    8.256    &    8.379    &    8.285    &    8.361    \\
\texttt{ambush\_5}    &    12.651    &    21.172    &    1.714    &    7.017    &    \textbf{1.639}    &    \textbf{6.411}    &    30.435    &    29.505    &    5.558    &    6.920    &    \textbf{5.262}    &    \textbf{6.386}    \\
\texttt{ambush\_6}    &    25.467    &    28.981    &    \textbf{4.409}    &    \textbf{7.535}    &    5.106    &    8.028    &    37.813    &    39.470    &    7.165    &    7.520    &    \textbf{7.083}    &    \textbf{7.457}    \\
\texttt{ambush\_7}    &    1.861    &    4.969    &    0.633    &    0.925    &    \textbf{0.584}    &    \textbf{0.891}    &    11.517    &    27.146    &    11.377    &    10.885    &    \textbf{11.270}    &    \textbf{10.763}    \\
\texttt{bamboo\_1}    &    1.104    &    1.221    &    \textbf{0.677}    &    \textbf{0.712}    &    0.677    &    0.713    &    8.159    &    8.560    &    \textbf{5.517}    &    \textbf{5.705}    &    5.517    &    5.713    \\
\texttt{bamboo\_2}    &    0.889    &    2.628    &    0.687    &    1.027    &    \textbf{0.686}    &    \textbf{1.014}    &    \textbf{7.134}    &    11.924    &    8.315    &    8.729    &    8.357    &    \textbf{8.669}    \\
\texttt{bandage\_1}    &    1.683    &    2.160    &    0.973    &    0.957    &    \textbf{0.936}    &    \textbf{0.935}    &    15.201    &    15.244    &    10.111    &    9.829    &    \textbf{9.862}    &    \textbf{9.681}    \\
\texttt{bandage\_2}    &    0.769    &    1.062    &    \textbf{0.329}    &    0.334    &    0.330    &    \textbf{0.331}    &    9.035    &    10.754    &    6.525    &    6.554    &    \textbf{6.523}    &    \textbf{6.542}    \\
\texttt{shaman\_2}    &    \textbf{0.286}    &    0.317    &    0.294    &    0.284    &    0.292    &    \textbf{0.284}    &    \textbf{5.571}    &    \textbf{5.643}    &    7.801    &    7.838    &    7.779    &    7.831    \\
\texttt{shaman\_3}    &    0.485    &    0.552    &    \textbf{0.178}    &    \textbf{0.291}    &    0.179    &    0.298    &    8.125    &    8.115    &    3.155    &    \textbf{4.559}    &    \textbf{3.153}    &    4.604    \\
\texttt{sleeping\_1}    &    0.502    &    \textbf{0.514}    &    0.149    &    0.541    &    \textbf{0.149}    &    0.541    &    5.939    &    6.110    &    1.954    &    \textbf{6.487}    &    \textbf{1.954}    &    6.488    \\
\texttt{sleeping\_2}    &    \textbf{0.146}    &    \textbf{0.143}    &    0.170    &    0.171    &    0.170    &    0.171    &    \textbf{2.508}    &    \textbf{2.520}    &    3.352    &    3.335    &    3.352    &    3.335    \\
\hline
\textbf{Overall}    &    0.701    &    4.122    &    0.488    &    1.379    &    \textbf{0.487}    &    \textbf{1.336}    &    7.899    &    13.009    &    6.826    &    7.213    &    \textbf{6.815}    &    \textbf{7.136}    \\
\hline
\end{tabular}
\vspace{-0.5em}
\end{table*}

We achieve the above by attaching a smoothly decaying kernel on each of our detected event points in $\textsc{Con}$ and $\textsc{Sep}$ and locally computing the weight for each event class.
Assuming a maximum radius of effect $\rho_e$ (free parameter) for our event points, we model the influence of each event with an RBF kernel of bandwidth $\sigma=\rho_e/3$.
The weights $w_f^i$ and $w_b^i$ of $\mathcal{W}_S^f$ and $\mathcal{W}_S^b$ for the source point $x_i^s$ are computed by accumulating influences of the event points in $\textsc{Con}=\{c_i\}$ and $\textsc{Sep}=\{s_i\}$ respectively that are within a $\rho_e$-radius from $x_i^s$:
\begin{align}
    w_f^i &= \frac{1}{Z}\left( 1 + \sum_{\mathclap{j\in\mathcal{N}_C(i)}} \exp\left(-\left\lVert x_i^s-c_j\right\rVert^2/(2\sigma^2)\right) \right)\textrm{, and}\label{eq:f_weight}\\
    w_b^i &= \frac{1}{Z}\left( \sum_{j\in\mathcal{N}_S(i)} \exp\left(-\left\lVert x_i^s-s_j\right\rVert^2/(2\sigma^2)\right) \right)\textrm{,}\label{eq:b_weight}
\end{align}
where $Z$ is a normalizing constant ensuring that $w_f^i+w_b^i=1$, and $\mathcal{N}_C(i)$, $\mathcal{N}_S(i)$ index the $\rho_e$-radius neighbors of $x_i^s$ in $\textsc{Con}$ and $\textsc{Sep}$ respectively.
In the absence of any topology event influence (e.g., for $x_i^s$ at least $\rho_e$ from any event point), the above defaults to $w_f^i=1$ and $w_b^i=0$, giving full weight to the standard forward hypothesis $\mathcal{W}_S^f$.
The local transformation of our final, topology-aware warp field estimate $\mathcal{W}_S=\{{T_S}_i,\ i=1,\ldots,\lvert S\rvert\}$ at location $x_i^s$ is then given by:
\begin{equation}
    {T_S}_i = \textrm{SE3}\left( w_f^i{T_S^f}_i + w_b^i{T_S^b}_i \right)\textrm{,}
    \label{eq:transform_blending}
\end{equation}
where $\textrm{SE3}(\cdot)$ converts the linear blend of the two transformation matrices back to a valid $SE(3)$ transformation matrix.
The complete blending process is summarized in Algorithm \ref{alg:local_hypothesis_blending} (see also equations \eqref{eq:forward_warp_1} and \eqref{eq:forward_warp_2}).
A visualization of the blending weights is given in Fig. \ref{fig:intermediate_bot} (middle), where each source geometry point is colored according to its inverted backward warp hypothesis weight $w_b^i$.

We note that, for source points close to topology events, one of $w_f^i$ and $w_b^i$ will dominate the other, effectively rendering the blending of \eqref{eq:transform_blending} a binary selection.
As we move farther from topology events, it is possible that the two weights assume comparable values (e.g., at points lying between two events of different type).
For our blended output \eqref{eq:transform_blending} to be seamless and error-free in that case, it is expected that $\mathcal{W}_S^f$ and $\mathcal{W}_S^b$ do not differ significantly in areas that are ``far enough'' from event points.
This highlights the importance of parameter $\rho_e$, which should be of adequate magnitude to cover event regions; we have found that values $\rho_e\ge 3r_b$ work well in practice, where $r_b$ is the resolution of our virtual deformation graph (Section \ref{sec:warp_field_optimization}).

As a concluding remark, we observe that the most costly operation of our post-processing phase is the calculation, based on radius-neighborhoods, of the stretch/compress values of Section \ref{sec:topology_event_detection}.
Algorithm \ref{alg:local_hypothesis_blending} also performs radius queries on point sets $\textsc{Con}$ and $\textsc{Sep}$, but the latter are typically very small in size.
In our experience, the overall running time of the entire phase is significantly smaller than a single warp field estimation.
Furthermore, in the case of RGB-D input, image structure can be easily exploited in order to accelerate the extraction of point neighborhoods.

\section{Experiments}\label{sec:experiments}
We conduct three sets of experiments for the evaluation of our registration pipeline.
The first one is performed on a public optical flow evaluation dataset and examines our algorithm's motion estimation accuracy, both with and without the topology handling phase (Section \ref{sec:motion_estimation_accuracy_evaluation}).
For the second one, we use a custom dataset with topology event annotations and evaluate our event detection performance, as well as our estimated warp field quality in the presence of separation events (Section \ref{sec:event_handling_evaluation}).
In our third set, we provide qualitative results in long-term model-to-frame registration scenarios on public RGB-D sequences that feature dynamic scene topology (Section \ref{sec:long_term_tracking}).

\subsection{General setup details}\label{sec:setup_details}
The input to all series of experiments is RGB-D data, either synthetic (first set) or captured by a Kinect-like RGB-D camera (second and third sets).

\textbf{Point cloud generation.} RGB-D frames are converted to point clouds equipped with surface normal and color information, as well as a sparse set of interest points derived from SIFT features.
In all cases, the full resolution of the input depth map is used, which is $1024\times 436$ for the synthetic sequences and $640\times 480$ for the camera data.
We use a fixed maximum depth of $5\si{\metre}$ for all sequences in the first set, and vary the cut-off value in the range of $0.8\si{\metre}$ to $2.0\si{\metre}$ for the camera data, depending on the sequence.
For normal estimation, we use $k$-NN neighborhoods with $k=30$ in our first set of experiments, and $\rho$-radius neighborhoods with $\rho=1.5\si{\centi\metre}$ in our second and third sets.
SIFT keypoints are extracted from the RGB images and lifted to 3D, discarding the ones that lie on depth boundaries.

\begin{figure}[!t]
    \centering
    \includegraphics[width=\columnwidth]{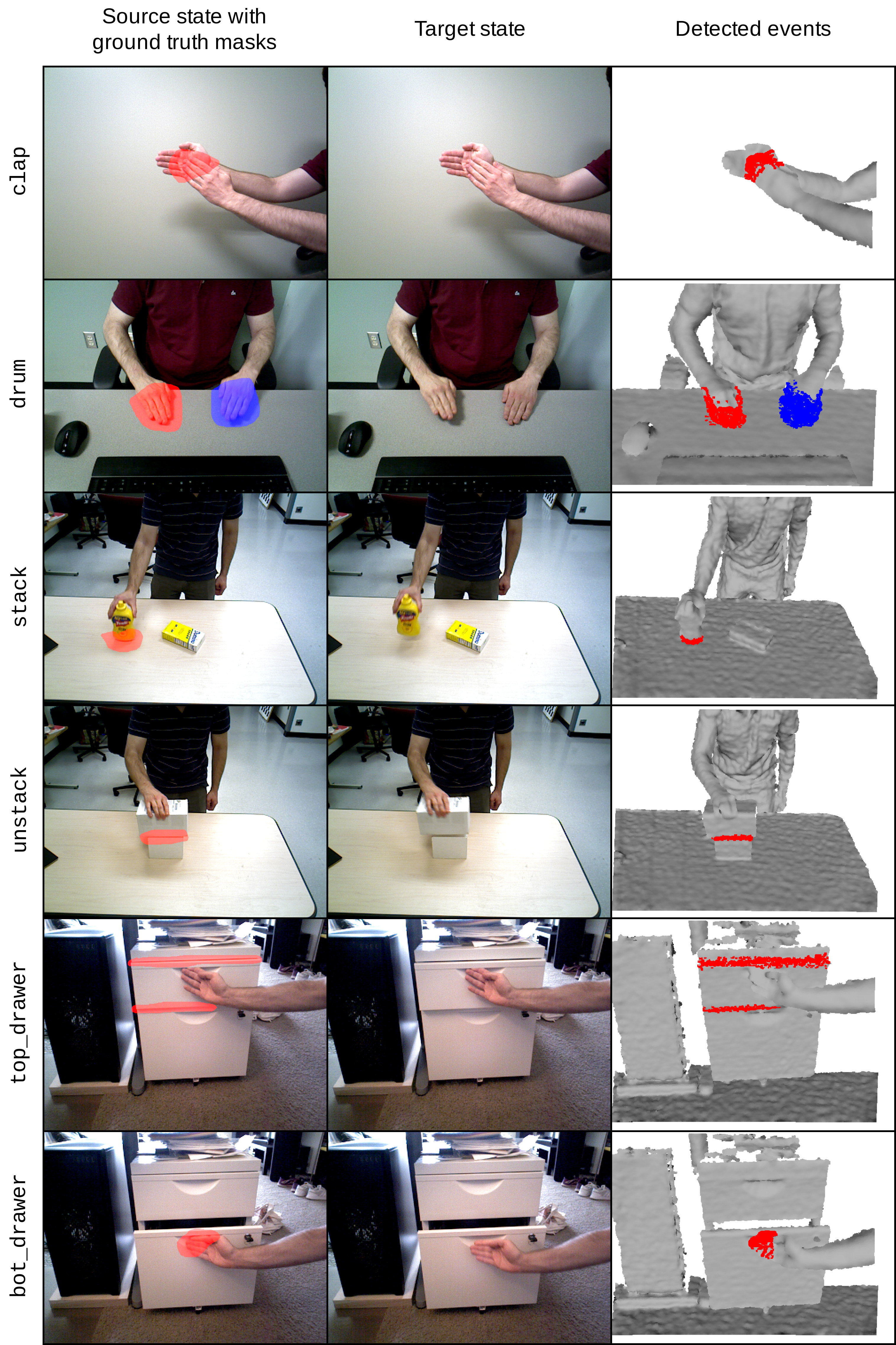}
    \vspace{-2.0em}
    \caption{Topological event detections on our dataset. Each row corresponds to a different sequence. First column: source color frame with event mask overlays (blue for contact, red for separation). Second column: target color frame. Third column: our topological event detections overlaid on the source geometry (blue for contact, red for separation).}
    \label{fig:topo_events}
    \vspace{-0.5em}
\end{figure}

\textbf{Warp field estimation.} In the correspondence association step (Section \ref{sec:correspondence_association}) of our non-rigid ICP algorithm, we set the maximum correspondence distance to $\theta_d=15\si{\centi\metre}$ for our first set of experiments and $\theta_d=5\si{\centi\metre}$ for the other sets, while we use common values $\theta_n=\ang{15}$ and $\theta_c=0.4$ for the maximum normal angle and color difference (colors are RGB triplets in $[0,1]^3$).
The embedded deformation graph $\mathcal{G}$ for our warp field parameterization (Section \ref{sec:warp_field_optimization}) has a resolution of $r_b=2.5\si{\centi\metre}$, with each node's area of effect being controlled by $\sigma_\textrm{def}=r_b/2$.
To evaluate the local transformation for each point in the source geometry (equation \eqref{eq:parameter_interpolation}), we use its 4 nearest neighbors in $\mathcal{G}$.
The point-to-point weight in our data term \eqref{eq:data_term} is set to $\lambda_\textrm{point}=2$.
To favor $L^1$-norm behavior by our regularization term \eqref{eq:regularization_term}, we use a small Huber loss parameter value of $\delta=10^{-4}$, while we set the term's weight to $\lambda_\textrm{stiff}=200$ (equation \eqref{eq:registration_objective}).
Regularization topology is given by the 6 nearest neighbor nodes of $\mathcal{G}$.
We perform a maximum of 10 top-level ICP iterations, while the process typically converges in less.
Within each optimization step, we perform a maximum of 5 Gauss-Newton iterations.

\textbf{Dynamic topology handling.} In all experiments, local stretch is computed on neighborhoods of radius $\rho_s=1.5\si{\centi\metre}$ (Section \ref{sec:topology_event_detection}).
To detect and classify topology change events, we use an asbsolute score threshold of $\tau=2.2$ and a relative ratio of $\alpha=1.5$.
For our blending step (Section \ref{sec:topology_hypothesis_blending}), we assume that every detected event has a radius of effect equal to $\rho_e=7.5\si{\centi\metre}$.

\subsection{Motion estimation accuracy evaluation}\label{sec:motion_estimation_accuracy_evaluation}
Due to the lack of publicly available datasets with ground truth dense 3D motion, we perform our accuracy assessments on MPI Sintel \cite{sintel}, a synthetic optical flow evaluation dataset.
The dataset contains multiple sequences of (typically) 50 frames that capture motions ranging from slow, almost rigid to very large, highly non-rigid ones.
In addition to ground truth optical flow, metric ground truth depth and camera intrinsics are provided, which we use to emulate RGB-D input.

We base our evaluation on two classical optical flow performance measures, the endpoint error (EPE) and the angular error (AE) \cite{baker2011database}.
If $\tilde{f}=(\tilde{u},\tilde{v})$ is an optical flow estimate at a given pixel whose ground truth value is $f=(u,v)$, EPE is computed as:
\begin{equation}
    e_\textrm{EPE}\left(\tilde{f},f\right)=\left\lVert\tilde{f}-f\right\rVert_2\textrm{.}
    \label{eq:epe_definition}
\end{equation}
The angular error AE is defined as the angle between the 3D space-time vectors $h(\tilde{f})=(\tilde{u},\tilde{v},1)$ and $h(f)=(u,v,1)$, as:
\begin{equation}
    e_\textrm{AE}\left(\tilde{f},f\right)=\arccos \frac{h(\tilde{f})^\top h(f)}{\lVert h(\tilde{f})\rVert_2 \lVert h(f)\rVert_2}\textrm{,}
    \label{eq:ae_definition}
\end{equation}
effectively enabling evaluation at pixels of zero flow.
We convert 3D motion estimates to optical flow by first warping the source points in 3D and then computing the 2D point/pixel displacements as differences of the projected endpoints onto the image plane.

% \begin{align}
%     e_\textrm{EPE} &= \left( (\tilde{u}-u)^2 + (\tilde{v}-v)^2 \right)^\frac{1}{2}\\
%     e_\textrm{AE}  &= \arccos\left( \frac{\tilde{u}u+\tilde{v}v+1}{\left( \tilde{u}^2 + \tilde{v}^2 + 1 \right)^\frac{1}{2} \left( u^2 + v^2 + 1 \right)^\frac{1}{2}} \right)
% \end{align}

We evaluate and compare three methods: PD-Flow \cite{jaimez2015primal}, a state-of-the-art scene flow algorithm, F-Warp, our general warp field estimation algorithm defined in Section \ref{sec:warp_field_estimation} (without the topology change handling phase), and FB-Warp, our complete topology-aware warp field estimation pipeline.
We run the three algorithms on the entire duration of 12 Sintel sequences and compute the average EPE and AE values per consecutive frame pair.
We report the median and mean frame-level average errors over each sequence in Table \ref{tab:sintel_evaluation}.
Median values are not easily affected by extreme values, often providing a better picture of how accurate estimation is ``half of the time''.
We also report overall mean and median error values for each method, computed over the total number of frames from all sequences.

\renewcommand{\arraystretch}{1.2}
\begin{table*}[!t]
\caption{Topology Change Event Detection}
\label{tab:topology_event_detection}
\vspace{-0.7em}
\centering
\begin{tabular}{| c | r r | r r r | r r r |}
\hline
\multirow{3}{*}{\textbf{Sequence}} & \multicolumn{2}{c|}{\textbf{Number of events}} & \multicolumn{3}{c|}{\begin{tabular}{@{}c@{}}\textbf{Ground truth to detected mapping}\\\textsc{GtToDet}\end{tabular}} & \multicolumn{3}{c|}{\begin{tabular}{@{}c@{}}\textbf{Detected to ground truth mapping}\\\textsc{DetToGt}\end{tabular}} \\
\cline{2-9}
\multicolumn{1}{|c|}{} & \multicolumn{1}{c}{\begin{tabular}{@{}c@{}}Ground\\truth\end{tabular}} & \multicolumn{1}{c|}{Detected} & \multicolumn{1}{c}{\begin{tabular}{@{}c@{}}Matched\\fraction \%\end{tabular}} & \multicolumn{1}{c}{\begin{tabular}{@{}c@{}}Mean\\overlap\end{tabular}} & \multicolumn{1}{c|}{\begin{tabular}{@{}c@{}}Mean\\delay\end{tabular}} & \multicolumn{1}{c}{\begin{tabular}{@{}c@{}}Matched\\fraction \%\end{tabular}} & \multicolumn{1}{c}{\begin{tabular}{@{}c@{}}Mean\\overlap\end{tabular}} & \multicolumn{1}{c|}{\begin{tabular}{@{}c@{}}Mean\\delay\end{tabular}} \\
\hline
\texttt{clap}    &    11    &    32    &    100.00    &    0.832    &    1.000    &    78.12    &    0.633    &    0.800 \\
 \texttt{drum}    &    21    &    63    &    100.00    &    0.763    &    -0.190    &    76.19    &    0.616    &    -0.042 \\
 \texttt{stack}    &    4    &    15    &    100.00    &    0.820    &    -0.250    &    40.00    &    0.713    &    0.000 \\
 \texttt{unstack}    &    4    &    20    &    100.00    &    0.689    &    -0.250    &    40.00    &    0.527    &    0.000 \\
 \texttt{separate}    &    5    &    17    &    100.00    &    0.661    &    -0.400    &    70.59    &    0.527    &    -0.583 \\
 \texttt{top\_drawer}    &    4    &    11    &    100.00    &    0.951    &    0.000    &    54.55    &    0.819    &    -0.667 \\
 \texttt{bot\_drawer}    &    3    &    12    &    100.00    &    0.882    &    0.000    &    66.67    &    0.757    &    -0.375 \\ \hline
 \textbf{Overall:}    &    52    &    170    &    100.00    &    0.788    &    0.058    &    66.47    &    0.630    &    0.035 \\
\hline
\end{tabular}
\vspace{-0.5em}
\end{table*}

Our FB-Warp method overall achieves the highest accuracy in terms of both error metrics, followed closely by our baseline, F-Warp.
PD-Flow is very accurate in estimating slow motions (e.g., in the \texttt{sleeping\_2} sequence), but falls behind in most cases, producing particularly large errors in sequences that contain very fast motions, such as \texttt{ambush\_5} and \texttt{ambush\_6}.
We also refer the reader to the Sintel-based evaluation of MC-Flow \cite{jaimez2015motion} (Table 2 of that paper),
another state-of-the-art scene flow algorithm with significantly better performance than PD-Flow in estimating large motions.
We were unable to evaluate MC-Flow ourselves, because its implementation has not been released.
While 6 of the Sintel sequences in our experiments and the ones in \cite{jaimez2015motion} are common, there are significant differences between our evaluation setups: \emph{1)} their reported EPE and AE values are computed on one specific frame pair per sequence, not over whole sequences as in here, \emph{2)} they downsample their input to half its original resolution per dimension (to $512\times 218$), whereas we use full resolution images, and \emph{3)} they consider only non-occluded pixels, while we compute errors on all valid ones.
The above prohibit reaching definitive conclusions.
However, the fact that we adopt an arguably more difficult evaluation strategy (at \emph{full resolution}, which means smaller pixel size for EPE interpretation, and evaluating over \emph{whole sequences}) and still obtain comparable EPE and AE absolute values to the ones reported in \cite{jaimez2015motion} (averages of 1.203 and 6.559, respectively), leads us to argue that FB-Warp compares favorably to MC-Flow.

\subsection{Dynamic topology event handling}\label{sec:event_handling_evaluation}
To evaluate the performance of the topology-handling phase of our pipeline, we collected, using a Kinect-like camera, an RGB-D dataset of 7 sequences that contain changes in scene topology.
The regions of visible topology changes were manually annotated on the color image as binary masks, drawn in freehand mode, and classified as either contacts or separations.
Our collected sequences capture the following diverse set of actions:
\begin{itemize}[noitemsep,topsep=0pt]
    \item Hand-clapping (\texttt{clap} sequence)
    \item Fast hand-drumming on a desk (\texttt{drum} sequence)
    \item Two pick-and-place actions on objects lying on a flat surface or on top of each other (\texttt{stack} and \texttt{unstack} sequences)
    \item A separation of two touching objects using both hands (\texttt{separate} sequence)
    \item Two drawer opening action sequences (\texttt{top\_drawer} and \texttt{bot\_drawer})
\end{itemize}
The annotated ground truth events capture all visible instances of hand-hand, hand-object, and object-object interaction.
Snapshots of our sequences are shown in Fig. \ref{fig:topo_events} (first two columns).

Our evaluation of this phase is twofold.
First, we assess how well our detected event points in $\textsc{Con}$ and $\textsc{Sep}$ (Section \ref{sec:topology_event_detection}) relate to the ground truth events.
Second, we qualitatively and quantitatively evaluate our topology-aware motion estimate (FB-Warp), including comparisons with our baseline algorithm (F-Warp), as well as two scene flow estimation methods.
% Second, we compare the geometric registration error between F-Warp (baseline warp) and FB-Warp (topology-aware motion estimate) in the vicinity of the ground truth topology change events, focusing on separations.

\subsubsection{Topology event detection}\label{sec:topology_event_detection_evaluation}

We uniformly represent topology change events as triplets $e_i=(l_i,t_i,X_i)$, where $l_i\in\{0,1\}$ is a binary label indicating contact or separation, $t_i\in\mathbb{N}$ is the time (frame index) of the event, and $X_i\subset\mathbb{R}^3$ is the subset of points in the source geometry that lie very close to event motion boundaries.
We use the superscript `gt' to denote ground truth event entities, and `det' to denote the ones associated with detections by our algorithm.
Let $\mathcal{E}^\textrm{gt}=\{ e_i^\textrm{gt} \}$ and $\mathcal{E}^\textrm{det}=\{ e_i^\textrm{det} \}$ denote the sets of ground truth and detected events, respectively.
Given an annotated sequence, these are populated according to the following:
\begin{itemize}[noitemsep,topsep=0pt]
    \item Labels and timestamps for events in $\mathcal{E}^\textrm{gt}$ come directly from the annotation data.
    Ground truth event point clouds $X_i^\textrm{gt}$ are obtained by the image annotation binary masks, which directly mask regions of the $t_i$-th frame's point cloud (input color and depth maps are registered).
    \item Detected events $\mathcal{E}^\textrm{det}$ are derived from the per-frame outputs of Algorithm \ref{alg:extract_topology_events}.
    At time $k$ (frame pair index), we interpret the \emph{connected components}, in the Euclidean sense, of point sets $\textsc{Con}$ and $\textsc{Sep}$ as separate, meaningful events.
    We insert all triplets $(0,k,\textsc{Con}_i)$ and $(1,k,\textsc{Sep}_i)$ into $\mathcal{E}^\textrm{det}$, where $\{\textsc{Con}_i\}$ and $\{\textsc{Sep}_i\}$ denote the respective connected component sets.
    % , and denote the respective connected component segmentations as $\{\textsc{Con}_i\}$ and $\{\textsc{Sep}_i\}$.
    % All triplets $(0,k,\textsc{Con}_i)$ and $(1,k,\textsc{Sep}_i)$ are inserted into $\mathcal{E}^\textrm{det}$.
    We use a distance threshold of $2\si{\centi\metre}$ for the Euclidean segmentation; to avoid noisy detections, we discard components that contain less than 75 points.
\end{itemize}
Adopting a 3D point set based representation for the spatial extent of topology events enables reasoning about event similarity in terms of metric distances.

Our assessments on spatial overlap of events will be based on the `$\rho$-overlap' metric, defined for a pair of point clouds $X_1$ and $X_2$ as:
\begin{equation}
    \textsc{Overlap}_\rho(X_1,X_2) \equiv \frac{\left\lvert S_\rho^{X_2}(X_1) \right\rvert + \left\lvert S_\rho^{X_1}(X_2) \right\rvert}{\lvert X_1\rvert + \lvert X_2\rvert}\textrm{,}
    \label{eq:overlap_metric}
\end{equation}
where $S_\rho^{B}(A)\subseteq A$ contains exactly the points in $A$ that lie within distance $\rho$ from their nearest neighbor in set $B$.
Clearly, $\textsc{Overlap}_\rho(X_1,X_2)\in[0,1]$.
It is easy to verify that this metric is simply the \emph{intersection-over-union} ratio for the sets $X_1\cup S_\rho^{X_1}(X_2)$ and $X_2\cup S_\rho^{X_2}(X_1)$.
We use a radius value of $\rho=3\si{\centi\metre}$.

\begin{figure*}[!t]
    \centering
    \includegraphics[width=\textwidth]{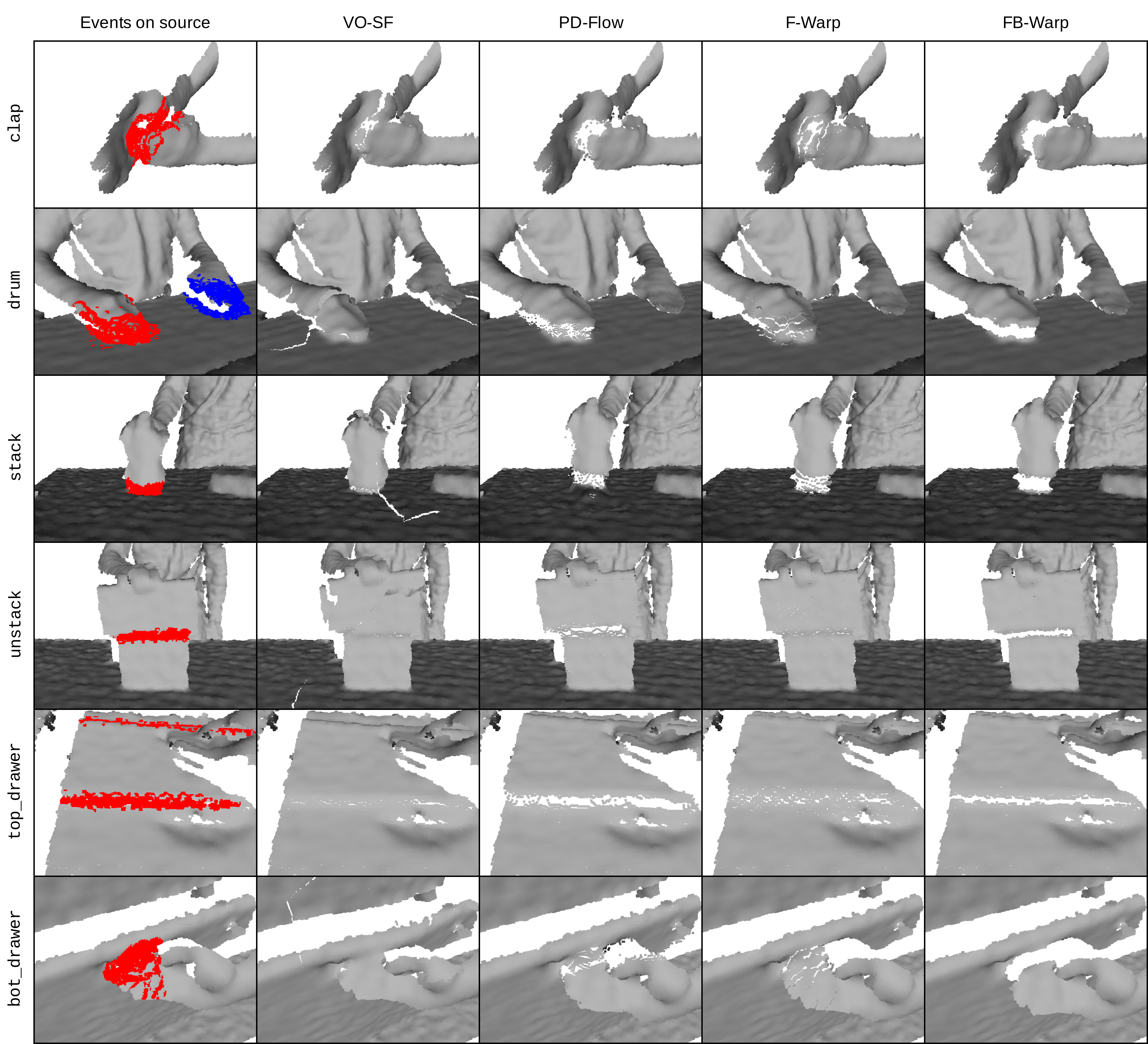}
    \vspace{-2.0em}
    \caption{Warping results on our dynamic topology dataset, with focus given to separation events. Each row corresponds to a different sequence. First column: our topological event detections overlaid on the source geometry (same as last column of Fig. \ref{fig:topo_events}, but rendered from a different viewpoint). Columns 2-5: warped source geometry under VO-SF, PD-Flow, F-Warp (our topology-agnostic baseline algorithm), and FB-Warp (our proposed approach).}
    \label{fig:topo_warp}
    \vspace{-0.5em}
\end{figure*}

We derive a many-to-many matching between sets $\mathcal{E}^\textrm{gt}$ and $\mathcal{E}^\textrm{det}$ by associating events in the two sets that have a significant spatiotemporal overlap.
A ground truth event $e_i^\textrm{gt}$ is matched to a detected event $e_j^\textrm{det}$ if and only if they both belong to the same class (contact or separation), their timestamps are very close, and they share a substantial spatial overlap.
If $\mathcal{M}=\{(i,j)\}$ is the set of event matches, then $\mathcal{M}$ contains all pairs $(i,j)$, for $i=1,\ldots,\lvert\mathcal{E}^\textrm{gt}\rvert$ and $j=1,\ldots,\lvert\mathcal{E}^\textrm{det}\rvert$, that satisfy all three conditions:
\begin{itemize}[noitemsep,topsep=0pt]
    \item $l_i^\textrm{gt}=l_j^\textrm{det}$
    \item $\lvert t_i^\textrm{gt} - t_j^\textrm{det} \rvert \leq 2$
    \item $\textsc{Overlap}_\rho(X_i^\textrm{gt},X_j^\textrm{det})\ge 0.2$
\end{itemize}
Based on the set of all valid spatiotemporal matches $\mathcal{M}$, we derive two interesting event mappings: one that maps each ground truth event to a single detected one, and an `inverse' one that maps each detected event to a ground truth one.
Both our `ground truth to detected' and `detected to ground truth' mappings associate an event in the first set with its match in the second one that maximizes overlap:
\begin{align}
    \textsc{GtToDet}(i) &= \argmax_{j:(i,j)\in\mathcal{M}} \textsc{Overlap}_\rho(X_i^\textrm{gt},X_j^\textrm{det})\textrm{,}\\
    \textsc{DetToGt}(j) &= \argmax_{i:(i,j)\in\mathcal{M}} \textsc{Overlap}_\rho(X_i^\textrm{gt},X_j^\textrm{det})\textrm{.}
\end{align}
Of course, the above are only defined for events (ground truth/detected) that have valid matches, i.e. $(i,j)\in\mathcal{M}$.

\begin{figure*}[!ht]
    \centering
    \includegraphics[width=\textwidth]{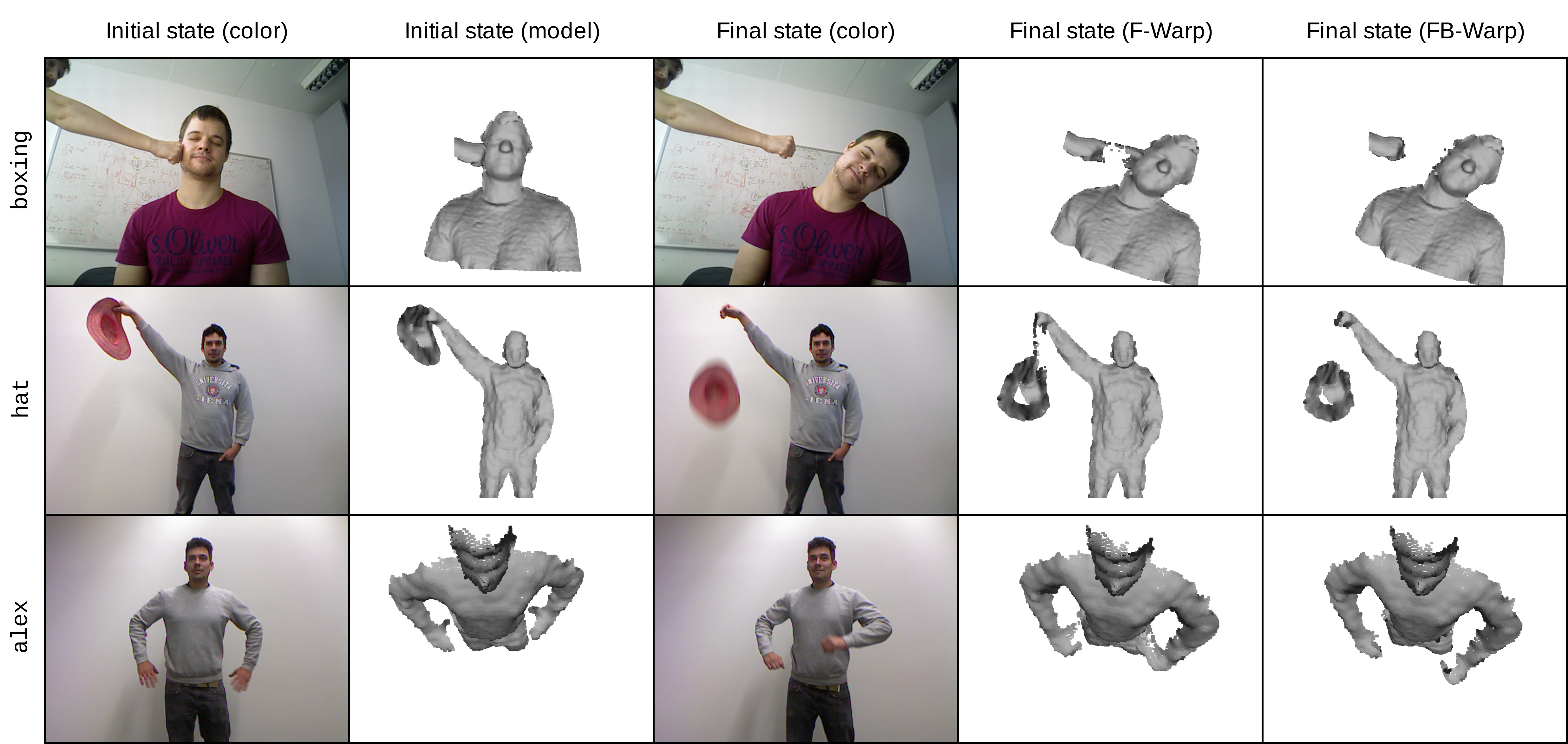}
    \vspace{-2.0em}
    \caption{Long-term model-to-frame registration. The \texttt{boxing} sequence (first row) is from \cite{innmann2016volume}. The \texttt{hat} and \texttt{alex} sequences (second and third rows) are from \cite{slavcheva2017killingfusion}. First column: color image at initial scene state. Second column: scene model in its initial state. Third column: color image at final scene state. Fourth and fifth columns: scene model in its final state after continuous model-to-frame deformation by our F-Warp (baseline) and FB-Warp (proposed) algorithms, respectively. The models for the \texttt{alex} sequence are rendered from a different viewpoint to better visualize the artifacts introduced by F-Warp (see Section \ref{sec:long_term_tracking}).}
    \label{fig:long_term}
    \vspace{-0.5em}
\end{figure*}

We present our detection results on our custom dataset in Table \ref{tab:topology_event_detection}.
On average, our pipeline extracts three times more events than the annotated ones (columns 2-3).
This is normal, as topology changes may manifest gradually in continuous video sequences, while our annotation process treats them as being instantaneous.
In columns 4-6 and 7-9, we evaluate each of the mappings $\textsc{GtToDet}$ and $\textsc{DetToGt}$ in terms of event coverage (fraction of $\mathcal{E}^\textrm{gt}$ and $\mathcal{E}^\textrm{det}$, respectively, that was matched), average spatial overlap, and average detection delay (signed difference $t_i^\textrm{gt}-t_j^\textrm{det}$).
All ground truth events are covered by our detections (column 4), while an average of $66.47\%$ of the detected events have a valid ground truth match (column 7).
These coverage fractions directly correspond to \emph{recall} and \emph{precision}, yielding an F-score of $0.8$.
At the same time, the spatial overlap of matched events is high (almost $80\%$ on average for the covered ground truth events) and within the error margins of our freehand annotation, while average detection delays are very small.
A small number of our detections (mostly separations) are depicted in the third column of Fig. \ref{fig:topo_events}, with the respective ground truth events shown in the first column.

We note that, in our context of non-rigid registration, high recall is more important than high accuracy, because missing a topology change event is very likely to result in motion estimation errors.
At the same time, as discussed in the concluding remarks of Section \ref{sec:topology_event_detection}, a small number of false positive detections have essentially no impact on motion estimation under reasonable assumptions.
Therefore, our topology change detection mechanism has desirable properties from a motion estimation perspective, while, at the same time, being able to detect contacts and separations with reasonable accuracy.

\subsubsection{Registration under dynamic topology}\label{sec:registration_topology_evaluation}
We now evaluate our registration accuracy in the presence of dynamic topology.
As discussed before, object separation events tend to induce warp field artifacts when not accounted for, while standard `forward' warp estimates properly handle contacts.
Therefore, we focus our assessments on areas of separation events.
For each ground truth separation event, we compute the average point-to-point distance between points in the \emph{warped} source geometry and their nearest neighbors in the target frame.
More specifically, the source geometry is given by the RGB-D frame associated with the annotated separation event, while the target one is given by its next frame in the sequence.
In order to avoid obscuring the differences between the two estimates in the areas of interest, instead of averaging over the whole source (event) frame, we only consider points within our ground truth annotation masks, which provide reasonable approximations of the true motion boundaries.
Furthermore, we discard occluded warped points using a simple depth test against the target geometry's depth map, with a tolerance threshold of $\Delta z_\textrm{occ}=1\si{\centi\metre}$.

\renewcommand{\arraystretch}{1.2}
\begin{table}[!t]
\caption{Registration Under Close-To-Open Topology}
\label{tab:registration_dynamic_topology}
\vspace{-0.7em}
\centering
\begin{tabular}{| c | r r r r |}
\hline
\multirow{2}{*}{\textbf{Sequence}} & \multicolumn{4}{c|}{\textbf{Registration error} (in $\si{\milli\metre}$)} \\
\cline{2-5}
\multicolumn{1}{|c|}{} & \multicolumn{1}{c}{VO-SF} & \multicolumn{1}{c}{PD-Flow} & \multicolumn{1}{c}{F-Warp} & \multicolumn{1}{c|}{FB-Warp} \\
\hline
\texttt{clap}    &    4.652    &    1.155    &    1.326    &    \textbf{1.135}    \\
\texttt{drum}    &    2.678    &    \textbf{1.006}    &    1.608    &    1.207    \\
\texttt{stack}    &    3.209    &    1.565    &    1.996    &    \textbf{1.452}    \\
\texttt{unstack}    &    6.457    &    2.371    &    3.138    &    \textbf{2.192}    \\
\texttt{separate}    &    2.249    &    \textbf{1.735}    &    2.418    &    1.985    \\
\texttt{top\_drawer}    &    1.995    &    1.383    &    2.305    &    \textbf{1.325}    \\
\texttt{bot\_drawer}    &    5.680    &    1.773    &    2.380    &    \textbf{1.222}    \\ \hline
\textbf{Overall:}    &    3.846    &    1.570    &    2.167    &    \textbf{1.503} \\
\hline
\end{tabular}
\vspace{-0.5em}
\end{table}

In our comparisons, we include four different motion field estimation algorithms: VO-SF \cite{jaimez2017fast}, PD-Flow \cite{jaimez2015primal}, F-Warp (our baseline), and FB-Warp (our proposed method).
We report per-sequence average registration errors (in $\si{\milli\metre}$) over separation areas in Table \ref{tab:registration_dynamic_topology}.
FB-Warp is more accurate in most sequences, achieving an average error reduction of about $30\%$ over our F-Warp baseline, with PD-Flow being a very close second.
VO-SF produces significantly less accurate results, because of the coarse pre-segmentation step on which its piecewise-rigid model is based.
We also provide qualitative registration results for a subset of our ground truth separation events in Fig. \ref{fig:topo_warp}.
VO-SF introduces seam artifacts to the warped geometry, as a result of its pre-segmentation step.
As the latter is highly unlikely to align with separation boundaries, the algorithm does not preserve motion discontinuities.
F-Warp, as expected, significantly oversmooths separation motion boundaries.
FB-Warp and PD-Flow exhibit the best performance, with the former producing appreciably cleaner surface separations.

\subsection{Long-term model-to-frame registration}\label{sec:long_term_tracking}
In our last set of experiments, we qualitatively evaluate the behavior of our proposed approach in long-term non-rigid model-to-frame tracking scenarios.
As input, we use some of the public RGB-D videos that accompany the dynamic reconstruction systems \cite{innmann2016volume} (\texttt{boxing} sequence) and \cite{slavcheva2017killingfusion} (\texttt{hat} and \texttt{alex} sequences) and exhibit dynamic scene topology.
We focus our attention on sequence intervals that capture topological events.
In particular, we initialize and fix a scene model that captures the scene state shortly before the topological events manifest and then continuously warp it towards each subsequent RGB-D frame.
Every time step involves registering the scene model (in its deformed state that aligns it to the previous frame) to the current frame.
Other than model-to-frame alignment, we perform no other updates on the scene model throughout the interval duration (e.g., no points are added, removed, or fused with more recent observations).

We show the results of our continuous model-to-frame deformation during our time intervals of interest in Fig. \ref{fig:long_term}, where we compare the final state scene models produced by our F-Warp baseline and our FB-Warp proposed method.
The final states correspond to time points well after the topological events completely manifested.
In the \texttt{boxing} sequence, we focus on the hand-head separation event.
We initialize our scene model using the partial reconstruction provided by the dataset (fusion of the first 101 frames), in which the hand is in contact with the head, and track the model for the 51 subsequent frames.
In the \texttt{hat} sequence, we focus on the hand-hat separation when the actor drops the hat and track our initial model for 16 frames.
In the \texttt{alex} sequence, we focus on a 27 frame interval towards the end of the sequence, where the actor first touches his torso with both hands (2 contacts) and then lifts them (2 separations).
For both \texttt{hat} and \texttt{alex}, the scene model was initialized from a single RGB-D frame at the beginning of our interval of interest.

As can be seen in the last columns of Fig. \ref{fig:long_term}, FB-Warp properly preserves motion discontinuities and introduces virtually no visible artifacts in the warped model geometry over the separation boundaries.
At the same time, the oversmoothing behavior of F-Warp is clearly visible, as objects tend to remain connected and there are no clean separations.
Standard non-rigid registration modules used in current dynamic reconstruction systems are expected to demonstrate a behavior similar to our F-Warp baseline (e.g., \cite{newcombe2015dynamicfusion}) or potentially introduce even more severe warping artifacts if they rely on quadratic loss regularization (e.g., \cite{innmann2016volume,dou2016acm,gao2018surfelwarp}).
On that note, although we have performed no relevant experiments, we believe that our proposed registration pipeline would significantly simplify building dynamic reconstruction systems, as it effectively suppresses the need to perform laborious regularization graph maintenance (initialization and updates), as in \cite{newcombe2015dynamicfusion} and \cite{innmann2016volume}, or handle dynamic topology in a post-registration stage that discards problematic regions and and reinitializes model tracking, as in \cite{dou2016acm} and \cite{gao2018surfelwarp}.

\section{Conclusions}\label{sec:conclusions}
We presented a complete pipeline for the non-rigid registration of arbitrary, unorganized point clouds that may be topologically different.
Building upon a general warp field estimation algorithm, we introduced an efficient topology event handling post-processing phase that detects and classifies object contact and separation events, and, by exploiting the different qualities of forward and backward motion estimates with respect to different event types, locally selects the most appropriate one, in a seamless manner.
% by exploiting the different qualities of forward and backward motion estimates in a seamless manner.
We evaluated the motion estimation accuracy of our method on the MPI Sintel dataset, achieving state-of-the-art performance.
Our evaluation on a custom dataset with sequences of highly dynamic scene topology demonstrated the success of our method in estimating motion on topological event boundaries, and showed promising performance in event detection.
To the best of our knowledge, this is the first approach to handle dynamic topology in the context of raw point cloud registration.
Furthermore, we openly release the implementation of our baseline warp field estimation algorithm as part of our point cloud processing library \cite{cilantro}.

In this work, we focused on improving dense motion estimation on separation boundaries by reasoning about two specific types of dynamic topology: `open-to-close' and `close-to-open'. There exist, however, object interactions that induce different types of topological changes, which our method is not equipped to handle.
One such interesting example is the case of an object \emph{sliding} on its supporting surface.
In this case, while our deformation criteria might give us some hints regarding the problematic areas, our inverted backward estimate is expected to share similar oversmoothing properties as a standard, forward warp field.
We are currently investigating insights that would allow us to efficiently tackle those situations, ideally without attacking the more general and (possibly) much harder problem of joint motion estimation and motion segmentation.

\section*{Acknowledgments}
The support of Northrop Grumman Mission Systems University Research Program, of ONR under grant award N00014-17-1-2622, and the support of the National Science Foundation under grants BCS 1824198 and CNS 1544787 are greatly acknowledged.
% The support of ONR under grant award N00014-17-1-2622 and the support of the National Science Foundation under grants BCS 1824198 and CNS 1544787 are greatly acknowledged.

\ifCLASSOPTIONcaptionsoff
  \newpage
\fi

\bibliographystyle{IEEEtran}
\bibliography{references}

% \newpage

\vspace{-2em}

\begin{IEEEbiography}[{\includegraphics[width=1in,height=1.25in,clip,keepaspectratio]{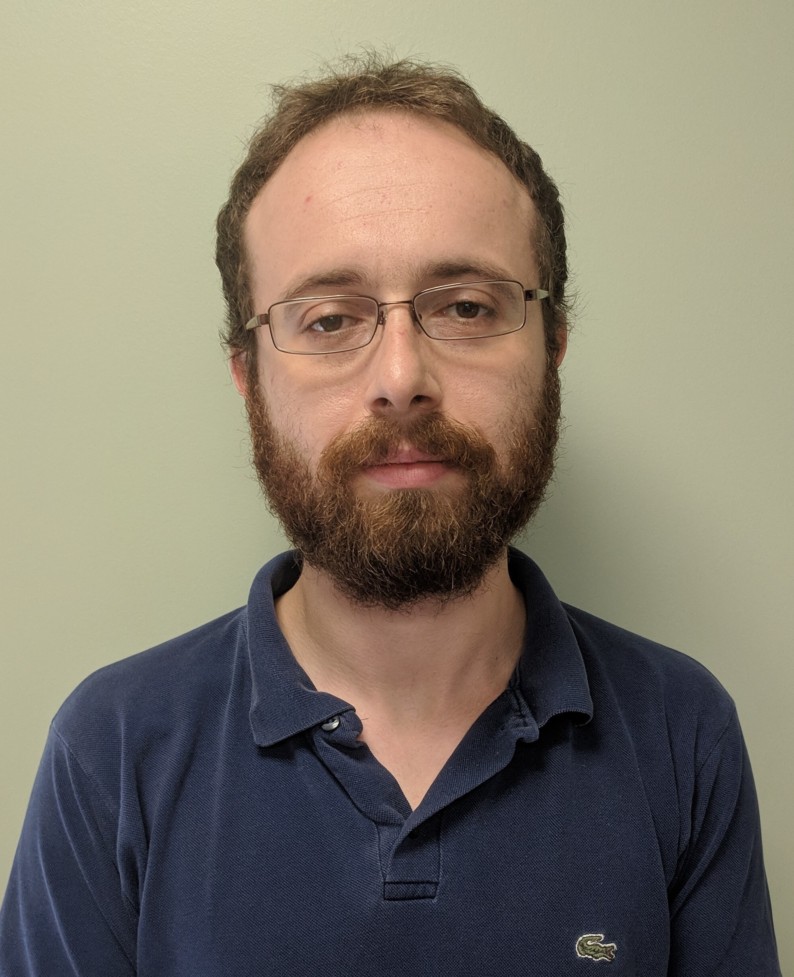}}]{Konstantinos Zampogiannis}
is a PhD candidate in Computer Science at the University of Maryland, College Park. He is working in 3D vision and action representations under the supervision of Prof. Yiannis Aloimonos and Dr. Cornelia Ferm\"uller. He completed his undergraduate degree at the National Technical University of Athens, Greece, under the supervision of Prof. Petros Maragos.
\end{IEEEbiography}

\vspace{-2em}

\begin{IEEEbiography}[{\includegraphics[width=1in,height=1.25in,clip,keepaspectratio]{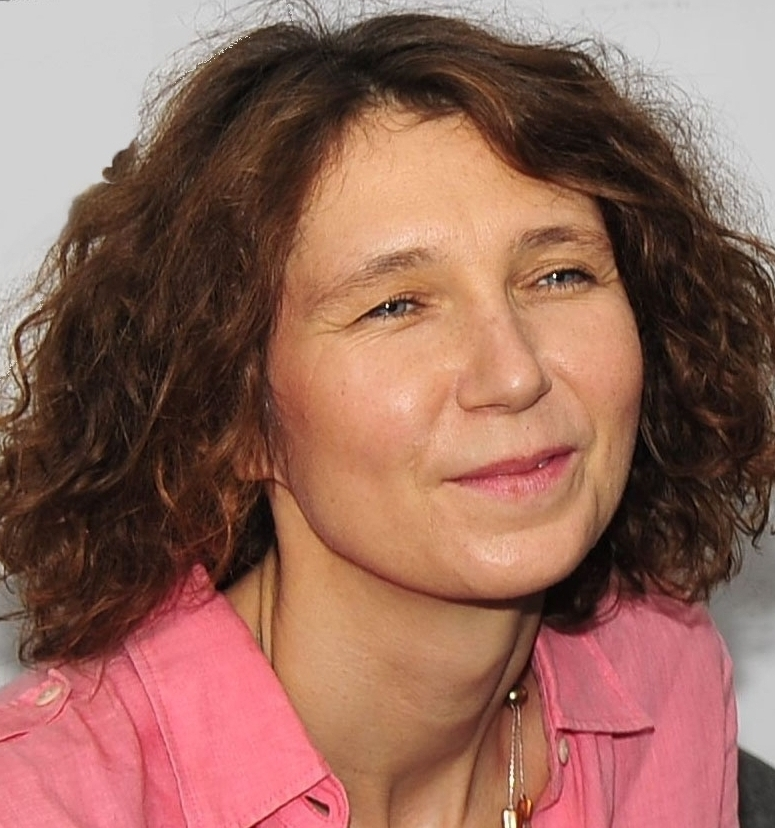}}]{Cornelia Ferm\"uller}
is a Research Scientist at the University of Maryland Institute for Advanced Computer Studies. She holds a Ph.D. from the Vienna University of Technology, Austria (1993) and an M.S. from the Graz University of Technology (1989), both in Applied Mathematics. Her research interest has been to understand principles of active vision systems and develop biological-inspired methods, especially in the area of motion. Her recent work has focused on human action interpretation and the development of event-based motion algorithms.
\end{IEEEbiography}

\vspace{-2em}

\begin{IEEEbiography}[{\includegraphics[width=1in,height=1.25in,clip,keepaspectratio]{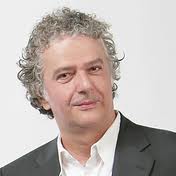}}]{Yiannis Aloimonos}
is Professor of Computational Vision and Intelligence at the Department of Computer Science, University of Maryland, College Park, and the Director of the Computer Vision Laboratory at the Institute for Advanced Computer Studies (UMIACS). He is also affiliated with the Institute for Systems Research and the Neural and Cognitive Science Program. He was born in Sparta, Greece and studied Mathematics in Athens and Computer Science at the University of Rochester, NY (PhD 1990). He is interested in Active Perception and the modeling of vision as an active, dynamic process for real time robotic systems. For the past five years he has been working on bridging signals and symbols, specifically on the relationship of vision to reasoning, action and language.
\end{IEEEbiography}

\vfill

\end{document}